\documentclass[letterpaper]{article} 
\usepackage{aaai2026}  
\usepackage{times}  
\usepackage{helvet}  
\usepackage{courier}  
\usepackage[hyphens]{url}  
\usepackage{graphicx} 
\usepackage{amsmath}
\usepackage{subfig} 
\usepackage{subfloat}
\usepackage{svg}
\svgsetup{
    inkscapeexe = your_path/inkscape.exe, 
    inkscapelatex = false                 
}
\usepackage{float}  
\usepackage{booktabs}
\usepackage{multicol}
\usepackage{multirow}
\usepackage{amsfonts}
\usepackage{times}
\usepackage{helvet}
\usepackage{courier}
\usepackage{xcolor}
\usepackage{natbib}  
\usepackage{caption} 
\usepackage{algorithm}
\usepackage{algorithmic}
\usepackage{appendix} 
\usepackage{newfloat}
\usepackage{listings}
\DeclareCaptionStyle{ruled}{labelfont=normalfont,labelsep=colon,strut=off} 
\floatstyle{ruled}
\newfloat{listing}{tb}{lst}{}
\floatname{listing}{Listing}
\usepackage{xcolor}
\usepackage{xspace}

\usepackage[utf8]{inputenc}

\newcommand{\thickhline}{%
    \noalign {\ifnum 0=`}\fi \hrule height 1pt
    \futurelet \reserved@a \@xhline
}

\DeclareRobustCommand\onedot{\futurelet\@let@token\@onedot}
\def\onedot{\ifx\@let@token.\else.\null\fi\xspace}

\def\vs{\emph{vs}\onedot}

\newcounter{corrcounter}
\nocopyright
\title{Real Garment Benchmark (RGBench): A Comprehensive  Benchmark for Robotic Garment Manipulation featuring a High-Fidelity Scalable Simulator}
\author{
    Wenkang Hu\textsuperscript{\rm 1}\equalcontrib,
    Xincheng Tang\textsuperscript{\rm 1}\equalcontrib, 
    Yanzhi E\textsuperscript{\rm 2},
    Yitong Li\textsuperscript{\rm 1},
    Zhengjie Shu\textsuperscript{\rm 1},
    Wei Li\textsuperscript{\rm 3}\thanks{Corresponding author(s).}\setcounter{corrcounter}{\value{footnote}},
    Huamin Wang\textsuperscript{\rm 2},
    Ruigang Yang\textsuperscript{\rm 1}\footnotemark[\value{corrcounter}]
}
\affiliations{
    \textsuperscript{\rm 1} Shanghai Jiao Tong University,
    \textsuperscript{\rm 2} Style3D Research,
    \textsuperscript{\rm 3} Nanjing University\\
}

\usepackage{bibentry}

\begin{document}

\maketitle

\begin{abstract}
While there has been significant progress to use simulated data to learn robotic manipulation of rigid objects, applying its success to deformable objects has been hindered by the lack of both deformable object models and realistic non-rigid body simulators. In this paper, we present \emph{Real Garment Benchmark} (RGBench), a comprehensive benchmark for robotic manipulation of garments. It features a diverse set of over 6000 garment mesh models, a new high-performance simulator, and a comprehensive protocol to evaluate garment simulation quality with carefully measured real garment dynamics. Our experiments demonstrate that our simulator outperforms currently available cloth simulators by a large margin, reducing simulation error by 20\% while maintaining a speed of 3 times faster. We will publicly release RGBench to accelerate future research in robotic garment manipulation. Website: \url{https://rgbench.github.io/}



\end{abstract}


\section{Introduction}




Robotic manipulation of deformable objects—most notably the challenge of handling garments—stands as a critical frontier in robotics research, with wide-ranging applications in household assistance and industrial automation~\cite{sanchez2018robotic}. These two major challenges for handling garments are (a) the vast, virtually infinite-dimensional state space of garments makes their configuration difficult to represent; and (b) the highly non-linear, under-actuated thin-shell dynamics, which makes their behavior difficult to predict~\cite{zhang2024achieving}. The high dimensional property and dynamic complexity are further exacerbated by pervasive contact and self-collision, where the thin fabric's constant folding and sliding create an intricate and rapidly changing landscape of physical interactions.

To overcome these challenges while mitigating the costs and risks of real-world trial-and-error, researchers increasingly turn to physics simulators, which offer a safe and scalable environment for developing manipulation policies. However, prevailing robotic simulators suffer from two critical limitations. First, their physical fidelity is often insufficient. To achieve the necessary computational performance, they rely on simplified models like Position-Based Dynamics (PBD), which are approximations of true continuum mechanics. As a result, they exhibit unrealistic physical behaviors, including distortions, stretching, and frequent self-penetration failures, creating a significant sim-to-real gap. Second, even with these simplifications, the performance of these simulators is often too slow. The computational resources required for large-scale parallel training of modern reinforcement learning algorithms are often unattainable. 

\begin{figure}[t] 
  \centering
  \includegraphics[width=0.9\columnwidth]{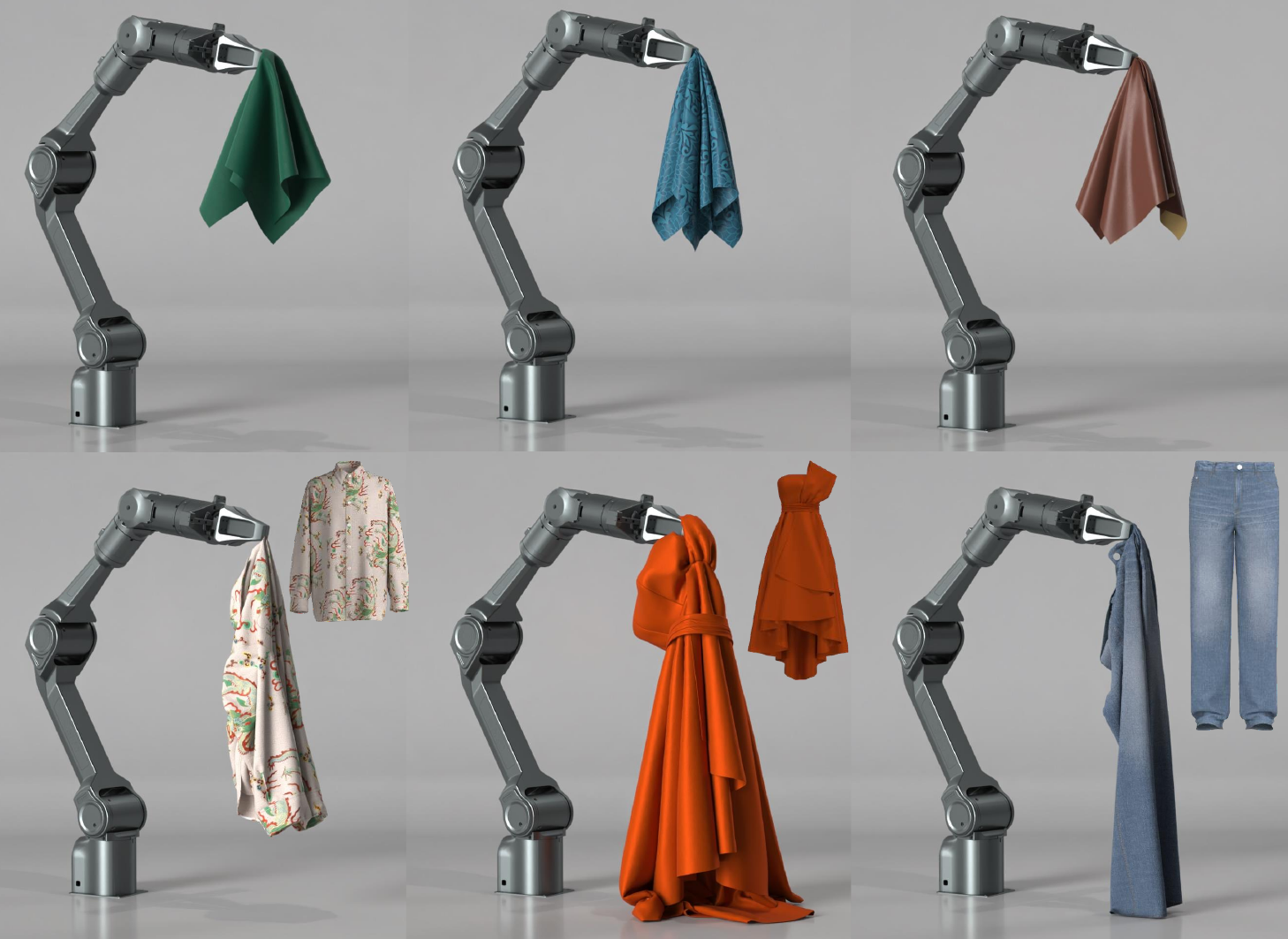} 
  \caption{Robotic Manipulation of Garments and Fabrics with Diverse Materials in RGBench} 
  \label{fig:example1}
\end{figure}

In order to address the sim-to-real gap, it is essential to have benchmarks and datasets that systematically evaluate the fidelity of simulation. The few existing benchmarks that attempt to quantify the sim-to-real gap are typically restricted to simple cases such as one-dimensional ropes~\cite{lim2022real2sim2real} or basic simple fabrics (e.g. handkerchiefs)~\cite{blanco2024benchmarking}. 

In this paper, we present Real-Garment Benchmark (RGBench) that includes \emph{GarmentDynamics},a novel high-speed and high-accuracy cloth simulator, and a diverse set of 3D garment models with physically accurate parameters. 
The contributions of this paper are:
\begin{itemize}
\item{
\textbf{A diverse 3D garment model dataset with measured physical properties and motions}. \hspace{0.1in} We introduce a novel public dataset for cloth manipulation. Its core feature is a rich collection of garments that span a wide variety of materials and topological complexities. Crucially, this dataset provides high-fidelity 3D ground-truth data for garment configurations resulting from both quasi-static and dynamic real-world robotic manipulations.}

\item{
\textbf{A novel cloth simulator with accuracy, robustness, and efficiency}.\hspace{0.1in}
We present {\it GarmentDynamics}, a novel physics-based simulator specifically engineered to overcome key limitations of existing robotics simulators in handling complex garment dynamics. Its design philosophy focuses on achieving physical accuracy, numerical stability, and collision robustness, and high computational performance through a combination of advanced physical models, accurate material property acquisition, and GPU acceleration. It will be released.}

\item{
\textbf{A dedicated benchmark for evaluating the sim-to-real gap of cloth simulators}.\hspace{0.1in}
We introduce Real-Garment Benchmark(RGBench), the benchmark designed to enable the rigorous evaluation and comparison of current mainstream physics simulators on the challenging task of garment manipulation. It provides a standardized framework to quantify the sim-to-real gap of any given simulator, featuring the richest collection of real-world garments and robotic actions to date.}

\end{itemize}

\section{Related Work}

\subsection{Deformable Object Manipulation Benchmark}
To systematically advance the field of robotic manipulation of deformable objects, benchmarks are essential to assess algorithm performance, define limitations, and establish standardized comparisons~\cite{longhini2024unfolding}. The majority of these evaluation platforms are situated within simulated environments, with mainstream robotics simulators like MuJoCo~\cite{todorov2012mujoco}, PyBullet~\cite{coumans2021}, and Isaac Sim~\cite{nvidia_isaac_sim} being common choices. Representative works include SoftGym~\cite{lin2021softgym}, and more recent GarmentLab~\cite{lu2024garmentlab} and DexGarmentLab~\cite{wang2025dexgarmentlab} were built in IssacSim, and Daxbench~\cite{chen2022daxbench}, which is based on a differentiable simulator. These benchmarks often utilize models from large open-source simulated datasets, such as ClothesNet~\cite{zhou2023clothesnet} and Cloth3D~\cite{bertiche2020cloth3d}, to create diverse evaluation scenarios.

Despite the sophistication of simulation-based benchmarks, the ``sim-to-real" gap remains a central challenge. Unlike rigid bodies that can be simulated with high fidelity, the extreme deformability of objects like cloth makes accurate modeling exceptionally difficult, as policies trained in simulation often fail when transferred to the real world. This has motivated a parallel line of research focused on curating real-world datasets, from large-scale image repositories like DeepFashion ~\cite{liu2016deepfashion}, household cloth collections with an RGB-D dataset from ~\cite{garcia2022household}. 

To benchmark the robotic manipulation for the deformable object,  more targeted efforts have sought to measure the reality gap. ~\cite{lim2022real2sim2real} quantified this gap for 1D cables in simulators like PyBullet and Isaac Sim. ~\cite{blanco2024benchmarking} evaluated the performance difference of various simulators in dynamic and quasi-static fling motions on simple cloth towels, chequered, and linens.
While these efforts represent valuable progress, their focus has largely been confined to structurally simple 1D cables and basic simple fabrics. A notable gap remains in the analysis of more complex garments, which we address by providing a benchmark with ground truth to measure the sim-to-real gap for common robotic clothing manipulation tasks: grasping, flinging, and folding.

\subsection{Physics-Based Cloth Simulation} 
A physics-based cloth simulator can represent cloth and its deformation in three main ways: as a network of springs~\cite{Choi:2002:SBR}, a collection of elements~\cite{Muller:2005:MDB, Volino:2009:ASA}, or a cluster of yarns~\cite{Kaldor:2008:SKC, Cirio:2014:YSW}. Among these, the element-based representation has become increasingly popular due to its balance between physical accuracy and computational cost.

Given a cloth representation, the key question is how to advance its dynamics over time. Explicit time integration~\cite{Bridson:2002:RTC, Bridson:2003:SCF} is conceptually simple, but requires very small time steps for stability, becoming computationally expensive. Implicit time integration~\cite{Baraff:1998:LSC} improves stability and allows larger time steps, but remains costly because it requires solving large linear systems.
Position-based dynamics (PBD)~\cite{Muller:2008:HPB, Muller:2014:SBD} was introduced as an alternative, emphasizing simplicity and robustness at the expense of physical accuracy and scalability, particularly for high-resolution meshes. Projective dynamics~\cite{Bouaziz:2014:PDF} later unified PBD and implicit integration under a common constraint-based framework, demonstrating their close relationship. This insight has inspired a line of research on fast cloth simulation~\cite{Wang:2015:CSA, Peng:2018:AAG, Chen:2024:VBD}, highlighting the potential of GPUs to accelerate implicit methods for real-time cloth simulation.


Efficient and robust collision handling remains one of the most challenging problems in cloth simulation. Over the years, researchers have explored various aspects of this problem — often leveraging GPUs — including collision culling~\cite{Tang:2011:VFC}, collision detection~\cite{Brochu:2012:EGE}, penetration untangling~\cite{Baraff:2003:UC}, and collision response methods~\cite{Tang:2016:CCM}. More recently, potential-based contact formulations~\cite{Li:2021:CIP, Wu:2020:ASF} have shown strong promise for robust collision handling, and recent work has focused on accelerating these methods on GPUs~\cite{Lan:2024:EGC, Li:2023:SPG}.

Beyond the core engine, realism depends critically on accurate physical parameters. These can be acquired through optimization-based methods that match observed motion \cite{Miguel:2012:DDE} or learning-based approaches that infer parameters directly \cite{Rasheed:2021:AVA}. However, the more fundamental distinction lies in how the measurement data is collected: either from videos of uncontrolled or unconstrained cloth motion \cite{Yang:2017:LBC}, or from carefully controlled experiments \cite{Wang:2011:DDE, Feng:2022:LBB}. We argue that controlled experiments are more reliable for isolating and precisely measuring individual material properties. For this reason, we also adopt strategies based on dedicated measurement devices.

\section{The RGBench Framework}

\begin{figure}[htbp] 
  \centering
  \includegraphics[width=1.0\columnwidth]{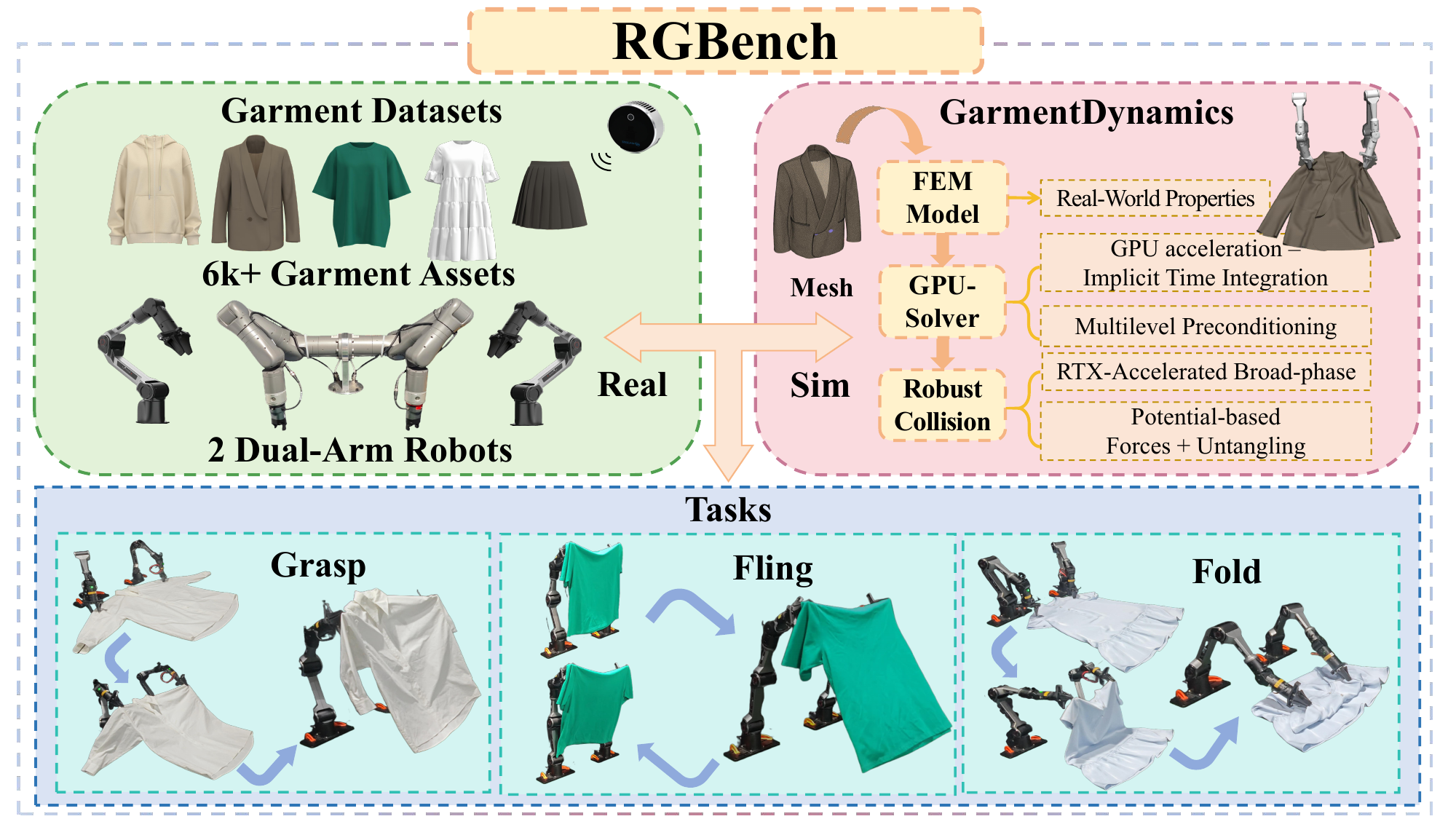} 
  \caption{Overview of RGBench framework}
  \label{fig:example2}
\end{figure}

Figure~\ref{fig:example2} presents an overview of the RGBench framework, which integrates a diverse garment dataset, dual-arm robotic setups, and the GarmentDynamics simulation system, covering three core tasks to bridge real-world and simulated garment interaction research.

\vspace{-0.3in}

\subsection{RGBench Dataset}

\subsubsection{Diverse Garment Assets}

\begin{figure}[htbp] 
  \centering
  \includegraphics[width=0.98\columnwidth]{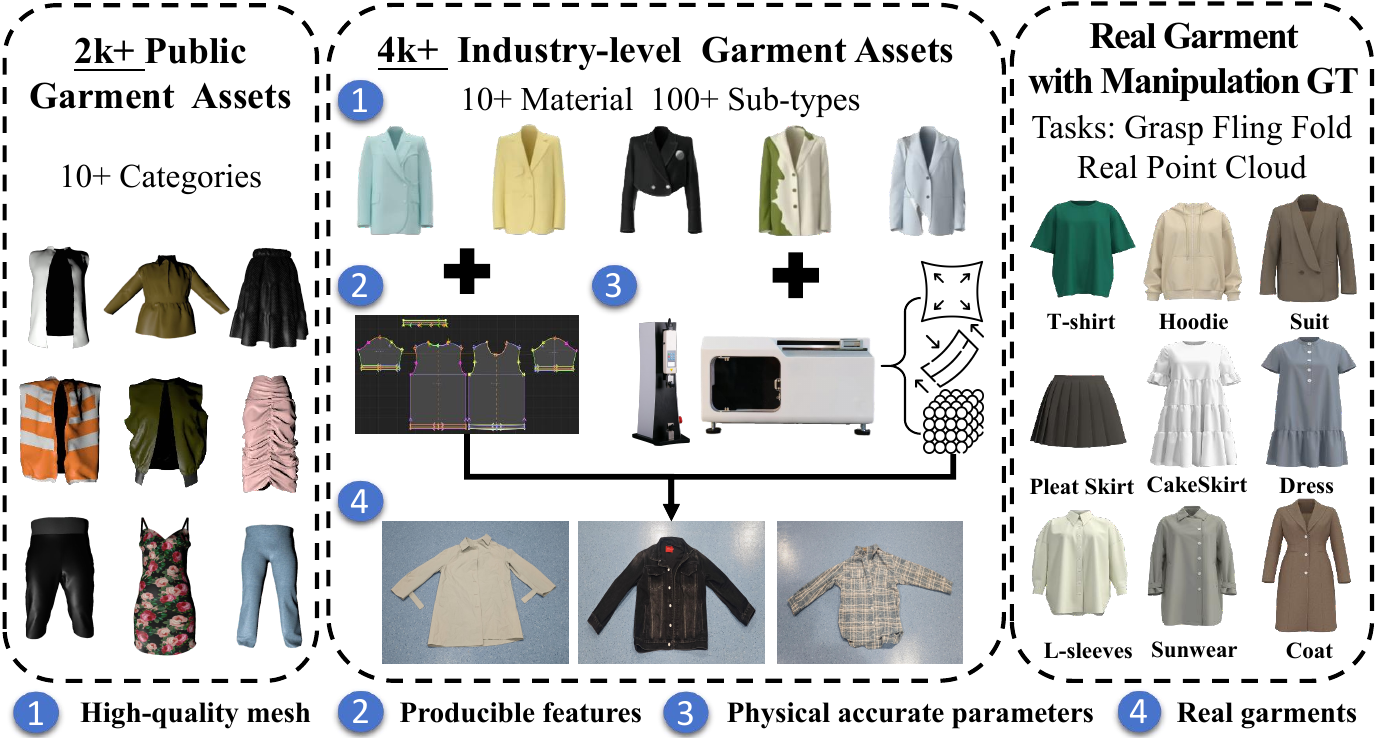} 
  \caption{RGBench garment dataset} 
  \label{fig:data}
\end{figure}

Our dataset encompasses a diverse range of garments with deliberate variations in size, style, and material to ensure comprehensive coverage of real-world manipulation scenarios shown in Figure~\ref{fig:data}. Specifically, the dataset encompasses a total of 6k+ garment models, consisting of two components: 4k+ self-collected industry-level production-ready assets, which feature high-quality meshes, producible features (including elasticity, folding edges, and seaming lines), and physically accurate parameters (such as stretch stiffness, bending stiffness, and density); and 2k+ garment meshes sourced from ClothesNet \cite{zhou2023clothesnet}. Specifically, the collection comprises garments of varying sizes (from small to large), diverse styles (fitted, loose-fitting, structured, and flowy), and a range of materials—cotton, linen, wool, polyester, nylon, silk, etc., each with distinct fabric properties. For manipulation tasks, we select 9 types of cloth to obtain the real point cloud during interaction as their ground truth (GT).

\vspace{-0.1in}

\subsubsection{Ground Truth Acquisition}
Our methodology for acquiring ground truth data is twofold, encompassing the static physical properties of each garment and its dynamic behavior during robotic manipulation.
\begin{itemize}
\item \textbf{Garment Physical Ground Truth}:
To create physically-grounded digital twins, we ensure both geometric and physical accuracy. For geometric accuracy, it is achieved by importing production-grade DXF pattern files to build 1:1 scale models, with seam lines defined to mirror the physical assembly. For physical accuracy, the Tensile Tester (SST1000) and Bending Tester (SBE1000) is used to profile the core mechanical properties of each fabric (e.g., tensile/bending stiffness, thickness, weight), as shown in part 3 by Figure~\ref{fig:data}. This measurement protocol adheres to established ASTM standards (D1388~\cite{astm_d1388_2018_misc}, D3107~\cite{astm_d3107_2017_misc}). This process yields a physically-grounded asset for simulation.

\item \textbf{Manipulation Ground Truth} We employ the dual-robotic setup (AGILEX Piper or JAKA K1 with DH PGC-50-35 grippers) to manipulate the garments. An Intel RealSense L515 camera records the entire process, capturing the complex deformations of the garment as a stream of RGB point cloud data, which are segmented by Grounded-SAM~\cite{ren2024grounded}.  This captured point cloud serves as the definitive ground truth ($P_{real}$) against which all simulation experiments are compared.

\end{itemize}

\subsection{RGBench Evaluation}
\subsubsection{Eval Tasks: Robotic Manipulation} 
Our benchmark is built on Grasp, Fling, and Fold—three foundational primitives that serve as the cornerstones of robotic garment manipulation. These primitives systematically address the core challenges of the field:  A robot must be able to reliably pick up a garment (Grasp), dynamically reconfigure its shape in the air (Fling), and precisely arrange it through contact-rich interactions (Fold). Each primitive is specifically designed to stress-test a distinct and critical aspect of the underlying physics simulation.

\begin{itemize}


\item \textbf{Grasp}:
The task begins with a flat garment, with predefined grasp points near the shoulder region. The robotic arms then synchronously grasp these points and lift the garment vertically. This primitive primarily evaluates the simulator's ability to model initial contact dynamics, frictional forces, and gravity-induced deformation.

\item \textbf{Fling}: 
Beginning with the garment held aloft by the grippers, the dual arms execute a rapid forward-then-backward trajectory to induce aerial fluttering of the garment. This task is specifically designed to challenge the simulator's handling of high-speed kinematics, inertial effects, aerodynamic drag, and large-scale deformation.

\item \textbf{Fold}: 
Starting again with a flat-laid garment,  the dual arms grasp the shoulder sections, lift slightly, and translate forward to the bottom edge. This procedure tests the simulator's capacity to handle complex, evolving self-contact, inter-surface friction, and the settling of the fabric into a stable, folded configuration.

\end{itemize}


\subsubsection{Eval Preprocess}
To ensure high-fidelity correspondence between our physical experiments and simulations, we perform a rigorous three-step alignment process. First, for initial state alignment, we standardize the garment's pose using a physical template of its outline. Second, the relative pose between the robot and camera is determined by hand-eye calibration or the Iterative Closest Point algorithm, which ensures high-precision registration. Finally, to achieve temporal synchronization, we apply a fixed, optimal time delay to the data streams to compensate for system latency.


\subsubsection{Eval Metrics}

We utilize Chamfer Distance(CD) and Hausdorff Distance(HD) as the core metrics for evalution. CD quantifies the average dissimilarity between two point sets. For a simulated mesh with vertices \(\mathcal{V}_t\) and a real-world point cloud \(\mathcal{P}_t\) at time t, the sim-to-real CD is defined as:
\begin{equation}
CD_{s2r}\left(\mathcal{V}_t, \mathcal{P}_t\right) := \frac{1}{|\mathcal{V}_t|} \sum_{v \in \mathcal{V}_t} \min_{p \in \mathcal{P}_t} \| v - p \|_1
\end{equation}
Conversely, the unidirectional real-to-sim CD reverses the point set order, calculated as:
\begin{equation}
CD_{r2s}\left(\mathcal{P}_t, \mathcal{V}_t\right) := \frac{1}{|\mathcal{P}_t|} \sum_{p \in \mathcal{P}_t} \min_{v \in \mathcal{V}_t} \| p - v \|_1
\end{equation}
where \(|\mathcal{V}_t|\) and \(|\mathcal{P}_t|\) denote the number of vertices in the corresponding mesh, and \(\| \cdot \|_1\) is the Manhattan distance. This dual formulation accounts for the disparity in point density between real and simulated data, ensuring both perspectives of alignment are captured.

HD focuses on capturing the worst-case alignment error. The unidirectional HD is:
\begin{equation}
HD_{s2r}\left(\mathcal{V}_t, \mathcal{P}_t\right) := \max_{v \in \mathcal{V}_t} \min_{p \in \mathcal{P}_t} \| v - p \|_1
\end{equation}
\begin{equation}
HD_{r2s}\left(\mathcal{P}_t, \mathcal{V}_t\right) := \max_{p \in \mathcal{P}_t} \min_{v \in \mathcal{V}_t} \| p - v \|_1
\end{equation}
These metrics highlight extreme deviations from both the simulated and real-world perspectives, which is crucial for identifying critical failures in dynamic cloth manipulation, such as large deformations or self-occlusions.

Unlike prior work using only sim-to-real metrics, we use both sim-to-real and real-to-sim metrics. Simulators, especially particle-based ones, have unstable garment simulations (e.g., model expansion/explosion) that inflate sim-to-real metrics, making them useless for comparison. Real-to-sim metrics show how well real garment surfaces match simulated meshes. We analyze both, focusing more on real-to-sim errors as they better reflect simulation fidelity to real observations and are less affected by simulator instabilities.





\subsection{Simulator}

To rigorously evaluate the simulation-to-reality gap, our benchmark is built upon an engine-agnostic framework. Any simulator integrated into this framework must support several key functionalities: the simulation of both rigid and deformable bodies, robust collision handling, variable integration timesteps, and the ability to query garment vertex states. To provide a comprehensive baseline, we incorporate three mainstream robotics simulators that meet these criteria: MuJoCo, PyBullet, and NVIDIA Isaac Sim.

\subsubsection{GarmentDynamics Design}
\paragraph{Core Components}

The following sections elucidate its architecture, detailing the design choices behind its GPU-accelerated implementation. We further analyze the mechanisms engineered to achieve superior accuracy, robustness, and computational efficiency for complex garment manipulation tasks.

\paragraph{Continuum-Based FEM Model}
In our simulator, cloth is treated as a continuous elastic surface discretized into a triangular mesh composed of linear triangular elements. The constitutive behavior is modeled in both in-plane and out-of-plane modes using an anisotropic extension of the model proposed in~\cite{Baraff:1998:LSC}, allowing us to account separately for the stretching and bending responses along the warp, weft, and diagonal directions of the fabric. Furthermore, we introduce nonlinearity into the out-of-plane behavior by defining the bending resistance as a quadratic function of the bending curvature. This formulation enables our model to more accurately capture cloth wrinkling, which strongly depends on bending resistance.

\paragraph{Physical Properties and Interactions}
The realism of a physics-based cloth simulator depends not only on the underlying physical models, but also on how accurately the material properties are measured and how effectively self-collisions and cloth–environment interactions are handled. Leveraging the physically grounded nature of our model, our simulator directly uses fabric properties measured from real-world samples, including mass density, thickness, elastic moduli, and surface roughness for friction. For collision handling, we combine potential-based contact forces with a collision untangling mechanism~\cite{Volino:2006:RSC}. This hybrid strategy achieves high performance while maintaining robustness, preventing penetrations even under frequent and severe contact conditions.

\paragraph{GPU-Accelerated Implicit Time Integration}
Finally, we advocate using an implicit time-integration scheme in our solver. Our solver leverages recent advances in GPU acceleration — such as multiresolution preconditioning~\cite{Wu:2022:AGB} and potential-based contact handling — to efficiently simulate cloth dynamics, even for meshes with a large number of degrees of freedom. Compared with the position-based dynamics used in Isaac Sim and explicit integration methods, implicit integration offers higher physical accuracy and improved numerical stability, even with large time steps and significant deformations.

\subsubsection{GarmentDynamics GPU Implementation}
\paragraph{GPU-Based Solver}
To achieve high performance, our simulator executes all computation stages entirely on the GPU, combining algorithmic advances with hardware-level optimizations. Cloth dynamics are solved using implicit Euler integration on the GPU, where each time step is formulated as an energy minimization of the following objective:
\begin{equation}
\mathcal{L}(\mathbf x) = \frac{1}{2 h^2}(\mathbf x - \mathbf x^t) \mathbf  M (\mathbf x - \mathbf x^t) + E(\mathbf x),
\label{eq:obj}
\end{equation}
where $h$ is the time step, $\mathbf{x} \in \mathbb{R}^{3N}$ denotes the unknown positional vector of the $N$ mesh vertices, $\mathbf{x}^t$ is the positional vector from the previous time step, $\mathbf{M} \in \mathbb{R}^{3N \times 3N}$ is the lumped mass matrix, and $E(\mathbf{x})$ represents the potential energy. Unlike~\cite{Baraff:1998:LSC}, our simulator solves Eq.~\ref{eq:obj} using a small number of inexact Newton iterations, each involving the solution of a linear system via a fixed number of preconditioned conjugate gradient (PCG) sub-iterations. Thanks to GPU acceleration, our solver achieves both fast convergence and efficient runtime.

\paragraph{Multilevel Preconditioning}
To further accelerate the convergence of our PCG solver, we adopt an algebraic multigrid preconditioner within the multilevel additive Schwarz (MAS) framework~\cite{Wu:2022:AGB}. This preconditioner is distinctive in that it approximates the system matrix with a block-diagonal inverse constructed from many small, non-overlapping subdomains across multiple resolution levels. At the start of the linear solve, the preconditioner computes the inverse of each block using Gauss–Jordan elimination and stores these inverses in GPU memory. During runtime, applying the preconditioner to a residual vector becomes a simple conflict-free, per-block sparse matrix–vector multiplication, with each block handled by a GPU thread. As demonstrated in~\cite{Wu:2022:AGB}, this preconditioner effectively reduces the condition number of the system, enabling the PCG solver to converge more rapidly within a fixed number of iterations.


 
\paragraph{Collision Detection}
Another major factor contributing to simulation performance is collision detection. To accelerate it, we develop a GPU-based bounding volume hierarchy structure, in which each query is efficiently executed in parallel using NVIDIA RTX ray intersection features. Compared with the general-purpose implementations in Isaac Sim or Bullet, our approach is significantly faster, capable of handling millions of broad-phase collision tests per second. Our implementation further employs optimized sparse-matrix data structures and CUDA kernel fusion to maximize memory coalescing and further boost raw throughput.

\section{Experiments and Results}

\textbf{Garment Manipulation Mode}:  Our framework provides two garment manipulation modes: robot interaction mode and pseudo interaction mode. (1) The robot mode utilizes URDF files and joint angles to simulate complete robotic behavior and garment dynamics simultaneously. (2) The pseudo mode achieves cloth manipulation simulation by directly controlling the movement of cloth intersection vertices, thereby simplifying the simulation of robotic arm motions, which focuses on the garment dynamics itself during manipulation.


\textbf{Deformable Parameters}:
Garment simulation involves a diverse set of parameters. For parameters with clear physical meanings that can be directly measured, we set their values based on experimental measurements using specialized instruments. For environmental parameters such as friction and damping, we first select a representative garment to perform parameter optimization. These optimized parameters are then applied uniformly across all garments, ensuring that environmental influences are consistent.

\subsection{Validation on a Foundational Cloth Benchmark}
To validate the fundamental accuracy of GarmentDynamics, we replicate the dynamic and quasi-static cloth deformation experiments on the rectangle chequered rag dataset from a widely accepted public benchmark (BCM) \cite{blanco2024benchmarking}. The garment state generated by GarmentDynamics is compared against the results from Mujoco and FLEX, which have the best performance in this dataset.

\begin{table}[h]
  \centering
  
  \resizebox{0.47\textwidth}{!}{ 
    \begin{tabular}{llccc}
      \toprule
      Mode & Metric & Mujoco & FLEX & Ours \\
      \midrule
      \multirow{2}{*}{Dynamic} & CD$_{s2r}$ & $0.067 \pm 0.026$ & $0.164 \pm 0.134$ & $\mathbf{0.062 \pm 0.028}$ \\
      & HD$_{s2r}$ & $0.154 \pm 0.035$ & $0.280 \pm 0.180$ & $\mathbf{0.150 \pm 0.051}$ \\
      \multirow{2}{*}{Quasi static} & CD$_{s2r}$ & $0.076 \pm 0.025$ & $0.072 \pm 0.019$ & $\mathbf{0.0389 \pm 0.006}$ \\
      & HD$_{s2r}$ & $0.186 \pm 0.055$ & $0.171 \pm 0.024$ & $\mathbf{0.094 \pm 0.022}$ \\
      \bottomrule
    \end{tabular}
  }
  \caption{Simulation results on BCM benchmark}
  \label{tab:ex1_results}
\end{table}

As shown in Table~\ref{tab:ex1_results}, GarmentDynamics demonstrates distinct strengths in quasi-static scenarios: it outperforms the state-of-the-art (FLEX) by \(\boldsymbol{46\%}\) in CD and \(\boldsymbol{45\%}\) in HD, validating its accuracy in capturing cloth deformations under slow, contact-dominated motions.

For dynamic tasks, we identified that a key factor affecting performance is the notable noise within the benchmark's input anchor point trajectories. To rigorously evaluate our simulator's resilience, we applied it directly to the raw data without the pre-processing steps. Despite that, GarmentDynamics still achieves the best performance, which highlights its superior robustness to imperfect inputs.


\begin{figure}[htbp] 
  \centering
  \includegraphics[width=0.98\columnwidth]{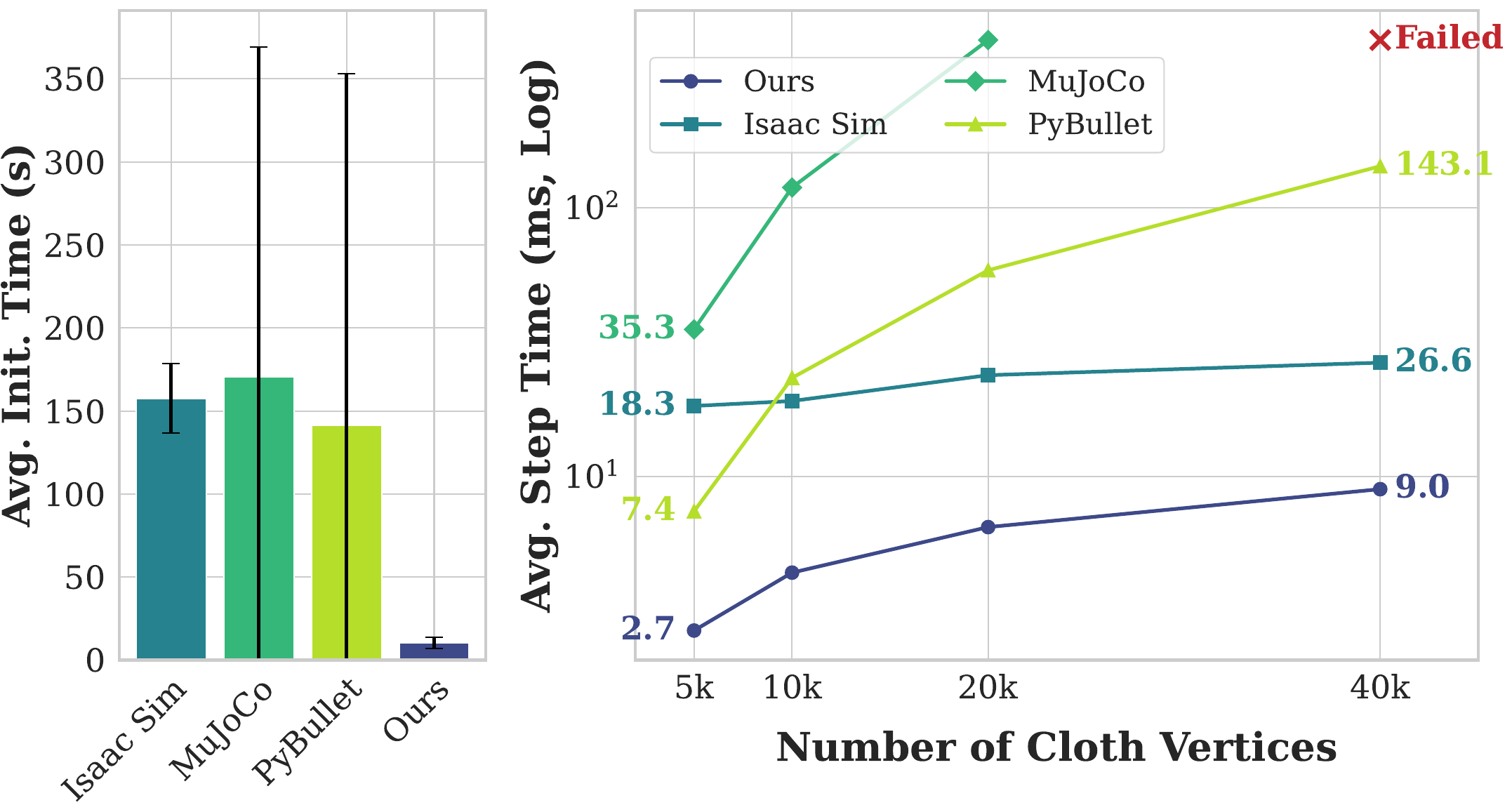} 
  \caption{\textbf{Simulator efficiency comparison}
   (left) Average Initial Time; (right) Average Step Time;}
  \label{fig:timme_efficient}
\end{figure}

\subsection{Simulation Efficiency and Scalability}

To evaluate the performance and scalability of our proposed simulator, we conduct a comprehensive benchmark against three widely used physics simulation environments: Isaac Sim, MuJoCo, and PyBullet. The benchmark task consisted of simulating a deformable cloth model with progressively increasing complexity, specifically with 5k, 10k, 20k, and 40k vertices. Two key performance metrics were measured: (1) Initialization Time, the one-time cost in seconds to load the scene and assets at 5k, 10k, 20k, and 40k vertices. (2) Average Step Time, the average computation time in milliseconds per simulation step, calculated over 100 iterations after the simulation reached a stable state.

The experimental results are shown in Figure 4, where our simulator shows a significant improvement in performance. Among the baseline simulators, IsaacSim is the most competitive performer, maintaining a respectable step time of 26.6 ms even at a high complexity of 40k vertices. However, its performance is less efficient at lower complexities. In contrast, our simulator consistently outperforms IsaacSim by  \textbf{3.0$\times$} to  \textbf{7.0$\times$}, achieving a minimum step time of just \textbf{2.7} ms. Conversely, both PyBullet and MuJoCo experience a dramatic increase in initialization and simulation time as the number of vertices grows.  Our simulator runs approximately \textbf{65.0$\times$} faster than MuJoCo at 20k vertices and \textbf{16.0$\times$} faster than PyBullet at 40k. Furthermore, MuJoCo fails entirely at the 40k vertex level due to computational complexity, highlighting the superior robustness of our approach. In addition to runtime execution, our simulator also exhibits superior initialization efficiency, reducing setup overhead by about \textbf{92\%} compared to PyBullet, and by over \textbf{93\%} on average compared to IsaacSim and MuJoCo.

\begin{table}[t]
\centering
\small
\setlength{\tabcolsep}{3.5pt}
\resizebox{1.0\columnwidth}{!}{
\begin{tabular}{l l | *{3}{c} | *{3}{c}}
\toprule
\multirow{2}{*}{Cloth} & \multirow{2}{*}{Action} & \multicolumn{3}{c|}{\textbf{CD$_{r2s}$}} & \multicolumn{3}{c}{\textbf{HD$_{r2s}$}} \\
\noalign{\vspace{1.5pt}}
\cline{3-8}
\noalign{\vspace{2pt}}
 &  & pybullet & isaacsim & ours & pybullet & isaacsim & ours \\
\midrule
Cakeskirt & Fling & 0.0588 & 0.0598 & \textbf{0.0406} & 0.1767 & 0.1872 & \textbf{0.1725} \\
Cakeskirt & Fold & 0.0266 & 0.0308 & \textbf{0.0179} & 0.1082 & 0.1082 & \textbf{0.0962} \\
Cakeskirt & Grasp & 0.0487 & 0.0517 & \textbf{0.0269} & 0.1573 & 0.1719 & \textbf{0.1222} \\
\midrule
Coat & Fling & 0.1071 & 0.0815 & \textbf{0.0379} & 0.2737 & 0.2406 & \textbf{0.1239} \\
Coat & Fold & 0.0309 & 0.0346 & \textbf{0.0279} & 0.1251 & 0.1182 & \textbf{0.1083} \\
Coat & Grasp & 0.0573 & 0.0605 & \textbf{0.0320} & 0.1817 & 0.1395 & \textbf{0.1053} \\
\midrule
Dress & Fling & 0.0889 & 0.0928 & \textbf{0.0687} & 0.2007 & 0.2228 & \textbf{0.1765} \\
Dress & Fold & 0.0331 & 0.0459 & \textbf{0.0187} & 0.1229 & 0.1338 & \textbf{0.0800} \\
Dress & Grasp & 0.0433 & 0.0490 & \textbf{0.0221} & 0.1468 & 0.1563 & \textbf{0.0953} \\
\midrule
Hoodie & Fling & 0.0310 & 0.0352 & \textbf{0.0256} & 0.1307 & 0.1493 & \textbf{0.1254} \\
Hoodie & Fold & 0.0240 & 0.0302 & \textbf{0.0225} & \textbf{0.0952} & 0.0966 & 0.1493 \\
Hoodie & Grasp & 0.0275 & 0.0308 & \textbf{0.0217} & 0.0947 & 0.1008 & \textbf{0.0882} \\
\midrule
Pleat Skirt & Fling & 0.0540 & 0.0388 & \textbf{0.0326} & 0.1236 & 0.1089 & \textbf{0.0736} \\
Pleat Skirt & Fold & 0.0256 & 0.0255 & \textbf{0.0126} & 0.1383 & 0.1199 & \textbf{0.0857} \\
Pleat Skirt & Grasp & 0.0241 & 0.0201 & \textbf{0.0159} & 0.0878 & 0.0908 & \textbf{0.0707} \\
\midrule
L-Sleeves & Fling & 0.0518 & 0.0600 & \textbf{0.0422} & 0.1770 & 0.2058 & \textbf{0.1547} \\
L-Sleeves & Fold & 0.0280 & 0.0308 & \textbf{0.0243} & 0.1252 & 0.1171 & \textbf{0.0982} \\
L-Sleeves & Grasp & 0.0347 & 0.0312 & \textbf{0.0280} & 0.1394 & 0.1123 & \textbf{0.1106} \\
\midrule
T-shirt & Fling & 0.0532 & 0.0567 & \textbf{0.0419} & 0.1522 & 0.1710 & \textbf{0.1197} \\
T-shirt & Fold & 0.0227 & 0.0314 & \textbf{0.0132} & 0.0778 & 0.0966 & \textbf{0.0512} \\
T-shirt & Grasp & 0.0348 & 0.0341 & \textbf{0.0226} & 0.1109 & 0.1058 & \textbf{0.0798} \\
\midrule
\bottomrule
\end{tabular}}
\caption{Metrics across garment types and actions in pseudo mode}
\label{tab:ours_last_column}
\end{table}

Experimental results confirm the superior computational efficiency of our proposed simulator. Compared to state-of-the-art simulators, our method exhibits significant advantages both algorithmically and in implementation. 
On the algorithmic side, we integrate multilevel preconditioning, inexact Newton iterations, and a hybrid collision-handling strategy. These techniques enable substantially faster convergence than the constraint-based methods employed by IsaacSim and PyBullet, particularly for high-resolution meshes with a large number of vertices. 
On the implementation side, our simulator leverages advanced GPU acceleration techniques, fully utilizing features such as RTX-based ray intersection and kernel fusion for optimal performance.

\subsection{Sim-to-Real Gap Across Manipulation Tasks}

To evaluate the consistency and robustness of GarmentDynamics in handling diverse physical interactions from quasi-static to dynamic modes, we design a standardized rubric of three core manipulation actions for each garment.

\begin{figure}[h] 
  \centering
  \includegraphics[width=1.0\columnwidth]{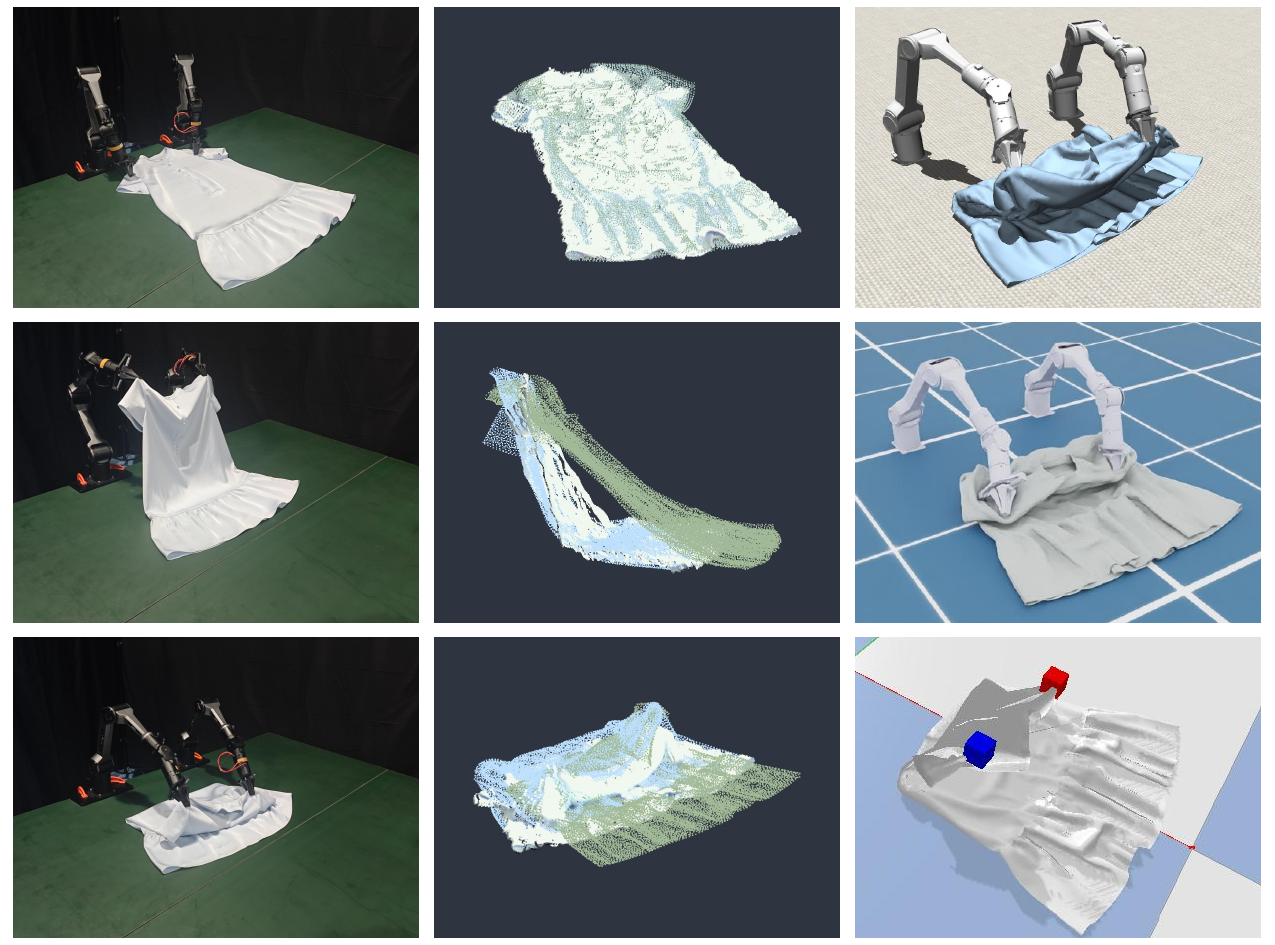} 
  \caption{\textbf{Real-to-Sim Comparison for Folding} (left column) Real-world robotic folding sequence.
  (middle column) Point cloud comparison: real-world (white), our simulator (blue), best of other simulators (Grey).
  (right column) Garment folding state in all simulators.} 
  \label{fig:simulation_compare}
\end{figure}

A qualitative comparison is present in Figure~\ref{fig:simulation_compare},  which visualizes the sim-to-real performance for folding a dress. The left column shows the real-world manipulation sequence; The center column visualizes the point-wise distances between the simulated and real point clouds; and the right column shows the process of three evaluated simulators. Notably, Isaac Sim exhibits instability upon gripper contact with the garment, often resulting in exaggerated twisting or inflation. PyBullet fails to generate valid grasping visuals, as its robot arm cannot reliably grasp the garment, only providing images of end-effector points. In contrast, GarmentDynamics remains stable throughout the entire sequence, owing to its robust hybrid collision-handling strategy. Most importantly, the point cloud generated by our simulator shows the closest alignment with real-world data, further validating its ability to replicate authentic garment behavior.

\begin{figure*}[t] 
  \centering
  \includegraphics[width=0.95\textwidth]{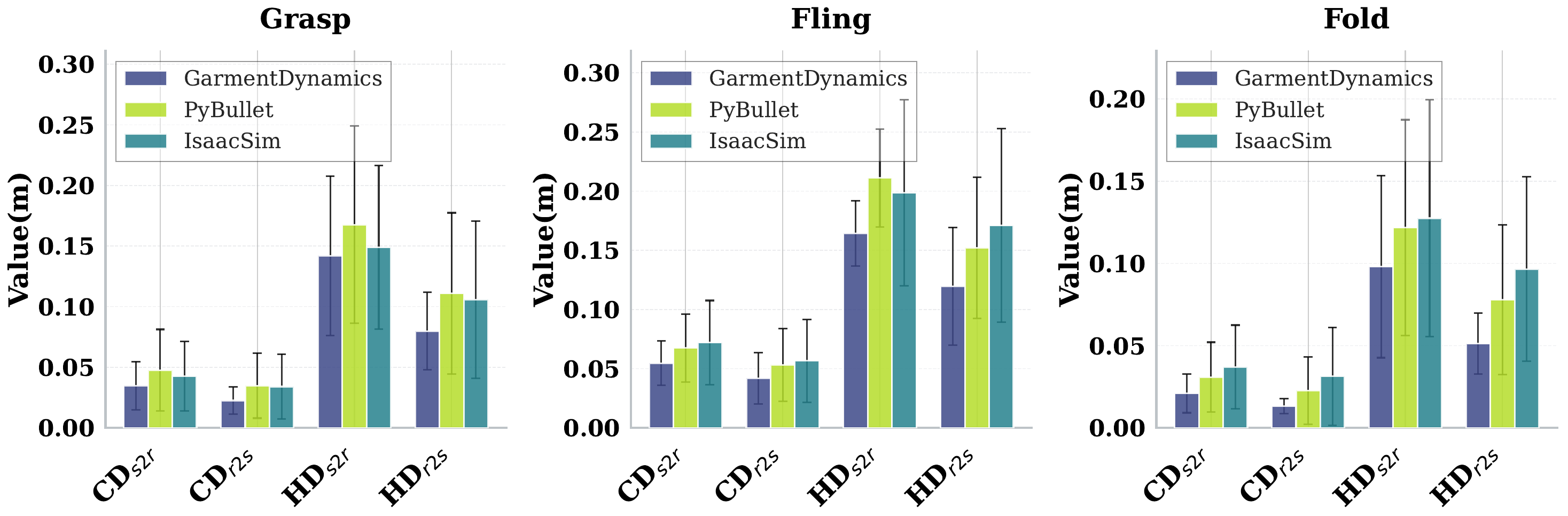} 
  \caption{Qualitative results of different actions for a T-shirt}
  \label{fig:ex3}
\end{figure*}

We extend this analysis quantitatively across the three manipulation tasks in Figure~\ref{fig:ex3}, using the T-shirt as a representative example.  Full results for all garments are included in the Appendix. Across tasks,  GarmentDynamics consistently achieves the smallest sim-to-real discrepancies, with narrow error bars highlighting its repeatability and stability. For grasp and fold tasks, GarmentDynamics exhibits a strong ability to accurately model quasi-static contact mechanics, where precise prediction of deformations is critical, with $CD_{r2s}$ error reduction up to 35\% and 58\%. Meanwhile, we observe that the sim-to-real gap widens for the dynamic manipulation task fling as the high-speed interaction. Despite this, GarmentDynamics still maintains a clear advantage in these dynamic scenarios, improving $CD_{r2s}$ and $HD_{r2s}$ by over 20\%. This versatility stems from its refined physical-based material modeling and robust GPU-based solver, which better accounts for real-world complexities like contact friction and inertial effects.

\subsection{Sim-to-Real Gap  Across Diverse Garments}


To further validate the superior material modeling capabilities and generalizability of GarmentDynamics, we expanded our experiments to include 7 distinct garment types in the pseudo and robot model. These garments span a wide range of variations in materials, structural designs, and topological complexities. The results are summarized in Table~\ref{tab:ours_last_column}.

\begin{figure}[t] 
  \centering
  \includegraphics[width=1.0\columnwidth]{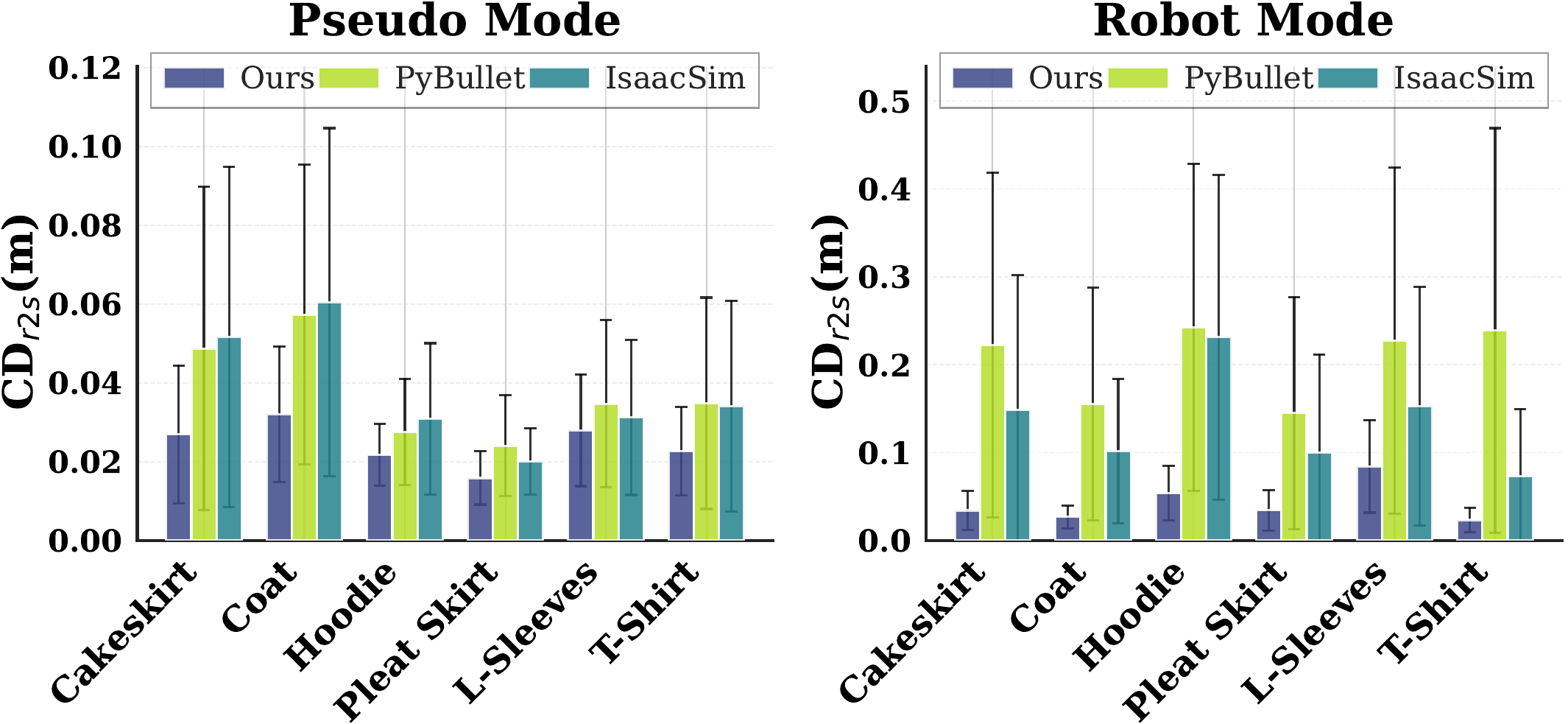} 
  \caption{Results of different garment types in grasp task} 
  \label{fig:ex4}
\end{figure}

GarmentDynamics demonstrates the lowest error across nearly all garment types and manipulation actions, thereby bridging the sim-to-real gap less than baseline simulators. To conduct a concrete comparison, we selected the grasp task as a representative to contrast how each simulator performs in Figure~\ref{fig:ex4}. Additionally, we also compare the two operation modes, providing a more comprehensive assessment of the simulators' behaviors in this experiment. We find the sim-to-real gap increases for thicker and longer garments, due to complex surface friction caused by spatial constraints. The result shows GarentDynamics achieves the lowest mean $CD_{r2s}$ errors, with consistent reductions of more than 37\% compared to PyBullet and Isaac Sim in pseudo mode. Its advantage is particularly pronounced for topologically complex items. For instance, in Cakeskirt manipulation, it reduces errors by up to 44\% in pseudo mode and 77\% in the robot mode — a substantial improvement that underscores its ability to handle intricate garment structures. 

\textbf{Manipulation Mode (Robot \vs Pseudo)}:
In robot mode, modeling inaccuracies can lead to slippage, insufficient grip strength, or erratic deformation, resulting in larger sim-to-real gaps compared to pseudo mode. PyBullet fails critically here: it struggles with failed grasps, severe penetration (grippers passing through garments), and overly thin cloth rendering, creating stark visual mismatches. These issues highlight challenges in modeling contact and material properties, while emphasizing our simulator’s superior fidelity in capturing physical interactions and visual realism.




\section{Conclusion and Discussion}

In this work, we addressed the dual challenge of the sim-to-real gap and performance limitations in robotic garment manipulation by Real-Garment Benchmark (RGBench), which includes a diverse real-world dataset and GarmentDynamics, a new high-fidelity, high-performance simulator. We followed the protocol defined in RGBench to evaluate performance across fundamental tasks such as grasping, flinging, and folding. The results clearly show that GarmentDynamics is the new state-of-the-art. It reduces the sim-to-real gap by over 20\% on average and by as much as 77\% for topologically complex garments. In terms of speed, it is 3.0x faster than its closest competitor, Isaac Sim, while robustly handling high-complexity scenes where others fail and slashing initialization time by over 90\%
Furthermore, we confirmed its fundamental accuracy on a prior benchmark for simple fabrics, where it surpassed the SOTA by over 45\%. 
In the future, we aim to further enhance the benchmark’s accuracy. Although perfect alignment of the garment’s initial wrinkle state remains challenging, improvements in sensor precision, delay compensation, and accurate measurement of friction and damping can reduce this gap. Looking ahead,  we will fully utilize the richness in RGBench to train a generalized action policy for garment manipulations. We also plan to enhance GarmentDynamics to handle more diverse conditions, such as wetness, and extend its application to more general deformable objects.





\bibliography{aaai2026}

\newpage

\appendix
\onecolumn
\part*{Supplementary Material}

\section{Appendix A \hspace{0.5em} Details of RGBench Datasets}
\vspace{0.5em}

\subsection{A.1 \hspace{0.5em} Diverse Types from Two Primary Sources} 

Self-Collected Industrial Assets (4k+ models): These are production-ready garments engineered for both simulation and physical fabrication. They cover a wide spectrum of everyday apparel types, such as casual shirts, formal dresses, loose-fitting sweaters, and structured blazers. Additionally, household textiles like absorbent dishcloths and decorative tablecloths are included to align with typical indoor manipulation scenarios.

ClothesNet Sourced Meshes (2k+ models): Leveraging the large-scale 3D garment dataset ClothesNet, we incorporate meshes from 11 carefully selected categories. These include Hat, Mask, Gloves, and Socks. This diverse range expands the coverage of specialized garment types for comprehensive manipulation research.

\subsection{A.2  \hspace{0.5em} Material Diversity and Physical Parameters} 

A core strength of the dataset is its emphasis on material variability—critical for sim-to-real transfer. We include 10 primary fabric types (cotton, linen, wool, polyester, nylon, silk, knit, velvet, leather, fur), each with distinct mechanical properties that significantly influence garment behavior during manipulation.

Key physical parameters are determined through rigorous processes.
Direct Measurements for Stretch and Bending Stiffness: For self-collected industrial assets, we utilize advanced tensile testing machines. These machines apply controlled forces to fabric samples, measuring the stretch stiffness and bending stiffness. For example, cotton fabrics, which are commonly used in casual wear, exhibit a high stretch stiffness of 1,000,000 \(g/s^2\), making them less prone to excessive stretching during grasping tasks compared to more elastic materials like knit. The area density (mass per unit area) is measured using precision weighing scales on carefully cut fabric swatches of known area. Thickness is gauged with specialized thickness gauges that ensure accurate measurement under standardized pressure. These parameters are calibrated against real-world samples to ensure that the simulation environment accurately mirrors the physical properties of the garments, which is crucial for tasks like folding where the thickness of the fabric affects the number of layers and the forces required.

Notably, the parameterization framework is designed for extensibility. Researchers can leverage the reference values in the table as a baseline and then independently adjust key parameters (stretch stiffness, bending stiffness, etc.) to simulate novel or hybrid fabric behaviors. For instance, by moderately reducing the stretch stiffness of a virtual cotton garment toward knit-like values, one can emulate the mechanical response of a cotton-knit blend. This flexibility supports the exploration of material generalization in manipulation scenarios, enabling the testing of algorithms under diverse, user-defined fabric properties that go beyond the physical samples in the dataset.

\begin{table}[h]
  \centering
  \resizebox{0.95\textwidth}{!}{ 
    \begin{tabular}{lccccc}
      \toprule
      Material & Sub - type Count & Stretch stiffness (g/s²) & Bending stiffness (g·m²/s²/rad) & Area Density (g/m²) & Thickness (mm) \\
      \midrule
      Cotton   & 22               & 1000000                   & 2800                             & 170                  & 0.26            \\
      Linen    & 10               & 130000                    & 500                              & 71                   & 0.19            \\
      Wool     & 11               & 380000                    & 650                              & 220                  & 0.38            \\
      Polyester& 17               & 100000                    & 1700                             & 109                  & 0.21            \\
      Nylon    & 5                & 65000                     & 100                              & 135                  & 0.32            \\
      Silk     & 10               & 30000                     & 150                              & 60                   & 0.19            \\
      Knit     & 20               & 25000                     & 200                              & 190                  & 0.5             \\
      Velvet   & 5                & 70000                     & 600                              & 210                  & 0.53            \\
      Leather  & 4                & 1000000                   & 20000                            & 500                  & 1.74            \\
      Fur      & 7                & 1700000                   & 2000                             & 230.3                & 0.67            \\
      \bottomrule
    \end{tabular}
  }
  \caption{Fabric material parameters}
  \label{tab:fabric_params}
\end{table}

\subsection{A.3 \hspace{0.5em} Industrial and Manufacturable Features}

\begin{figure*}[htbp] 
  \centering
  \includegraphics[width=0.95\textwidth]{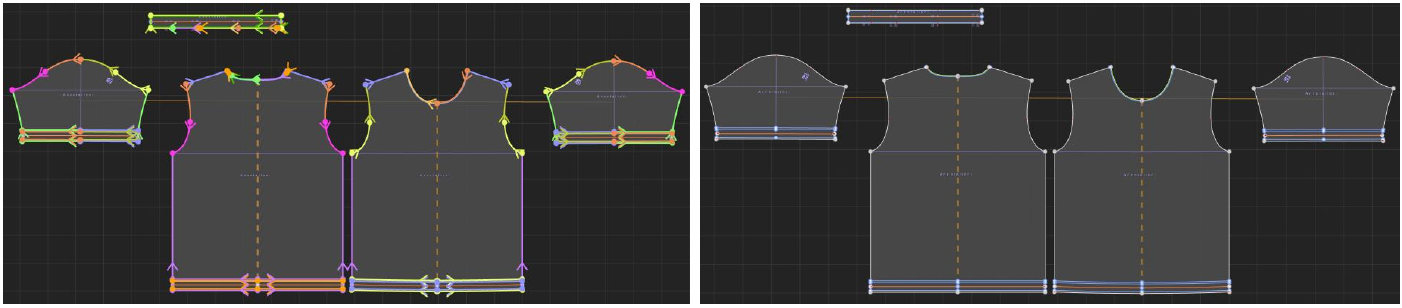} 
  \caption{Seaming line and feature edges}
  \label{fig:feature}
\end{figure*}

To bridge the gap between simulation and physical production, our dataset places a strong emphasis on producible features—design elements that are critical for the real-world fabrication and functional behavior of garments during manipulation. Figure~\ref{fig:feature} details the digital pattern design of a garment, showcasing two key components for physical fabrication and simulation. 

\begin{figure*}[htbp] 
  \centering
  \includegraphics[width=0.75\textwidth]{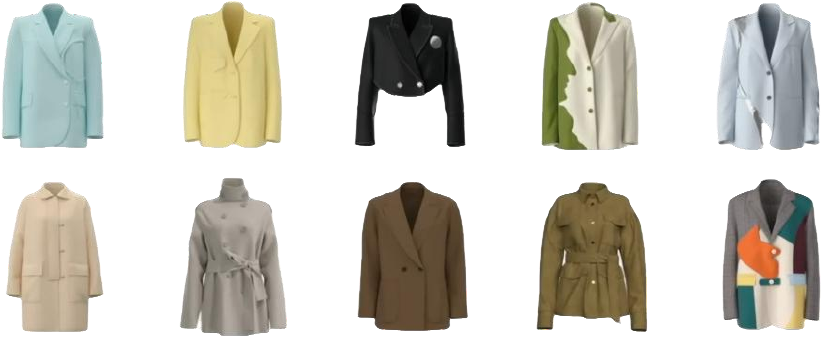} 
  \caption{High-fidelity meshes}
  \label{fig:mesh}
\end{figure*}

Color-Coded Pairs show the seaming lines. Identical colors mark corresponding edges (e.g., sleeve edges, shoulder seams) that will be stitched together. This ensures alignment during sewing, critical for structural integrity. These lines define how garment panels connect, directly influencing collision behavior and deformation in simulation—e.g., a stitched shoulder seam restricts stretching at that joint.
Yellow Lines (Elastic/Folding Edges) Represent zones where elastic materials are applied or intentional folding occurs. For example:
Elastic Edges: Govern stretchable regions (e.g., waistbands), modeled with parameters like “elastic length ratio” to simulate stretch/recovery.
Folding Edges: Define pre-set fold lines (e.g., hems), with folding angle controlling how sharply the fabric bends.
These edges introduce localized mechanical behaviors—elastic edges increase stretchability, folding edges reduce bending stiffness at specific zones—enabling accurate replication of real-world garment dynamics.
Standardized Topologies: The meshes of all garments in the dataset undergo thorough post-processing to ensure compatibility with simulation engines as shown in Figure~\ref{fig:mesh}. This includes creating watertight geometry, where all the edges and faces of the 3D mesh are properly connected without gaps or overlaps. This pre-processing reduces the need for researchers to perform extensive post-processing, allowing them to focus on developing and testing manipulation algorithms. 

Leveraging high-fidelity meshes, seaming lines, and feature edges, the garments within this dataset are production-ready, conforming to industrial-level standards. This ensures they can be directly utilized for physical fabrication, bridging the gap between simulated garment models and real-world manufacturing processes, and providing a robust resource for research in robotic garment manipulation.

\subsection{A.4 \hspace{0.5em} Ground Truth (GT) for Manipulation Tasks}

For core manipulation tasks (Grasp, Fling, Fold), we select 9 representative cloth types to capture real-world point clouds as GT. This ensures generalizability: By spanning materials from rigid (leather) to highly deformable (silk), the dataset supports robust algorithm development for diverse manipulation scenarios. A leather jacket, with its high bending stiffness and low deformability, presents a different challenge for robotic manipulation compared to a silk scarf, which is highly deformable and prone to complex folding patterns. Including such a wide range of materials ensures that the developed algorithms can handle the variability encountered in real-world garment manipulation.

\vspace{0.5em}
\section{Appendix B \hspace{0.5em} Details of Real-to-Sim Alignment}
\vspace{0.5em}
\subsection{B.1  \hspace{0.5em} World Coordinate and Origin Alignment}
\begin{figure*}[htbp] 
  \centering
  \includegraphics[width=0.7\textwidth]{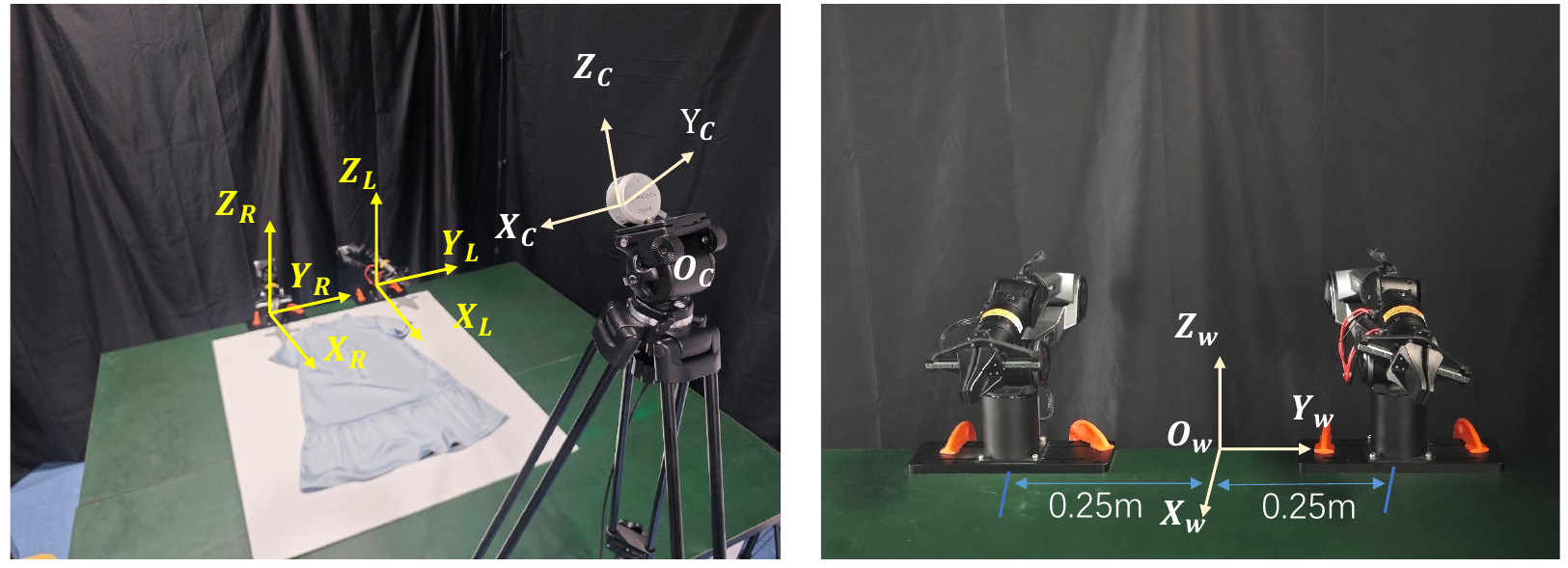} 
  \caption{Experiment Setup}
  \label{fig:exp_setup}
\end{figure*}

Our experimental setup is depicted in Figure~\ref{fig:exp_setup}. A camera is mounted approximately 2 meters from the workspace, angled to ensure a comprehensive view of the garment both on the tabletop and during manipulation.
The world coordinate system in our simulation is defined relative to the dual-arm Piper robot. The origin is set at the midpoint between the two robot bases. The coordinate system is right-handed,
with the +x axis pointing forward from the robot, the +z axis pointing upward, and the +y axis oriented along the line connecting the two arms.
The translation vectors for the left and right robot arm bases relative to this origin are $[0,\ -0.25,\ 0]$ and $[0,\ 0.25,\ 0]$m, respectively, each with an identity rotation.

\subsection{B.2 \hspace{0.5em} Garment Initial State Alignment}

\begin{figure*}[h] 
  \centering
  \includegraphics[width=0.7\textwidth]{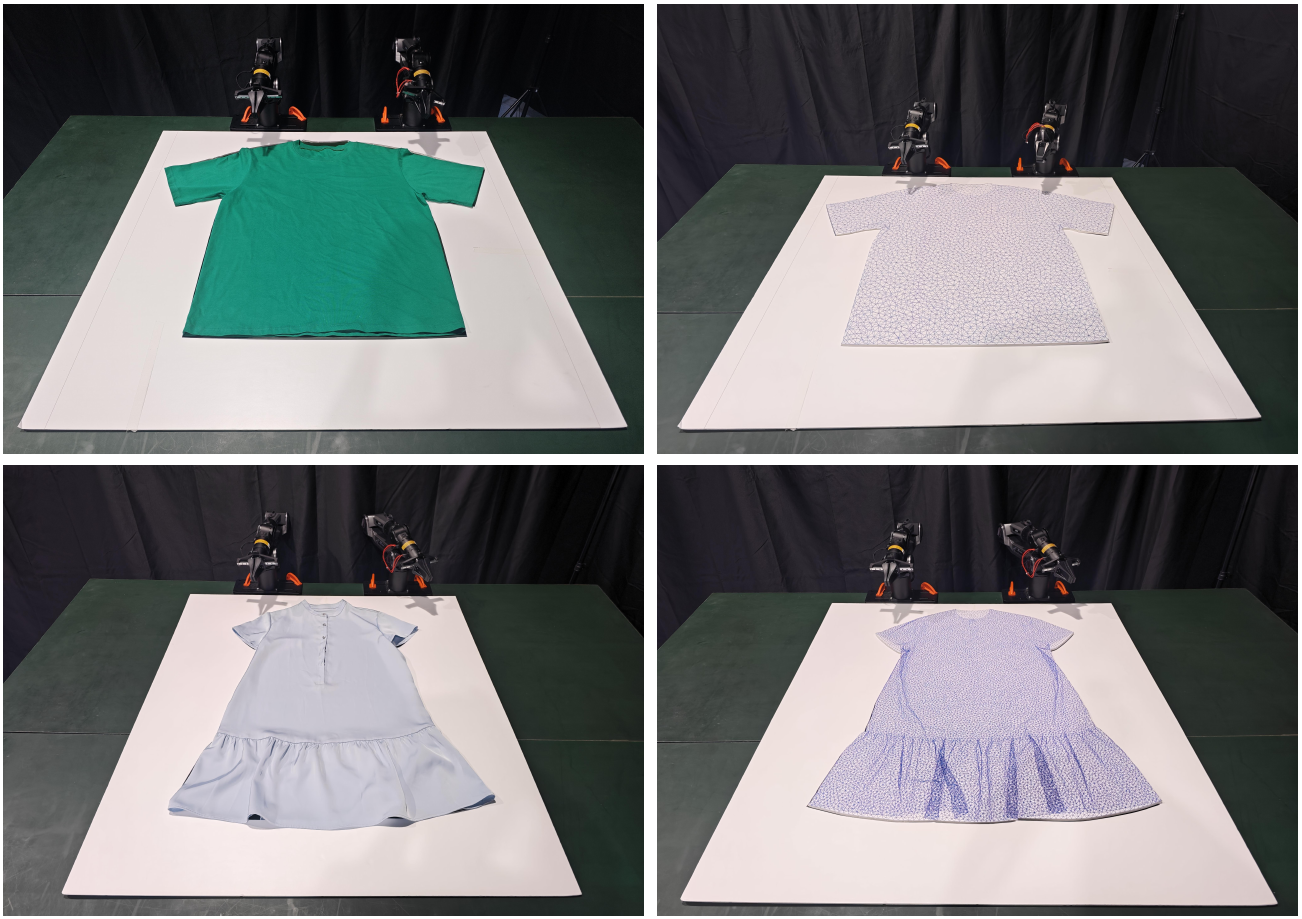} 
  \caption{Garment Initial State Template}
  \label{fig:initial_template}
\end{figure*}

To ensure consistent initial garment placement, we utilized a physical template, shown in Figure~\ref{fig:initial_template}, precision-cut to the garment's outline. This template standardizes the initial garment pose to a translation of [0.1,0.0,0.0] and an identity rotation. To prevent initial collision with the tabletop surface, the garment is initialized at a minimal height of 0.01 m. Thus, the complete initial state of the garment is represented by the 7D vector [0.1,0.0,0.01,1.0,0.0,0.0,0.0], corresponding to the position ($x$,$y$,$z$) and the orientation as a unit quaternion ($q_w$,$q_x$,$q_y$,$q_z$)


\subsection{B.3 \hspace{0.5em} Alignment of Camera Coordinate System and Robot Coordinate System}
We align the camera and robot coordinate systems using two complementary methods. The primary method is a standard hand-eye calibration. By affixing a checkerboard pattern to the robot's end-effector, we capture multiple configurations and solve the $AX=XB$ problem to find the rigid transformation between the camera and robot frames.

\begin{figure*}[htbp] 
  \centering
  \includegraphics[width=1.0\textwidth]{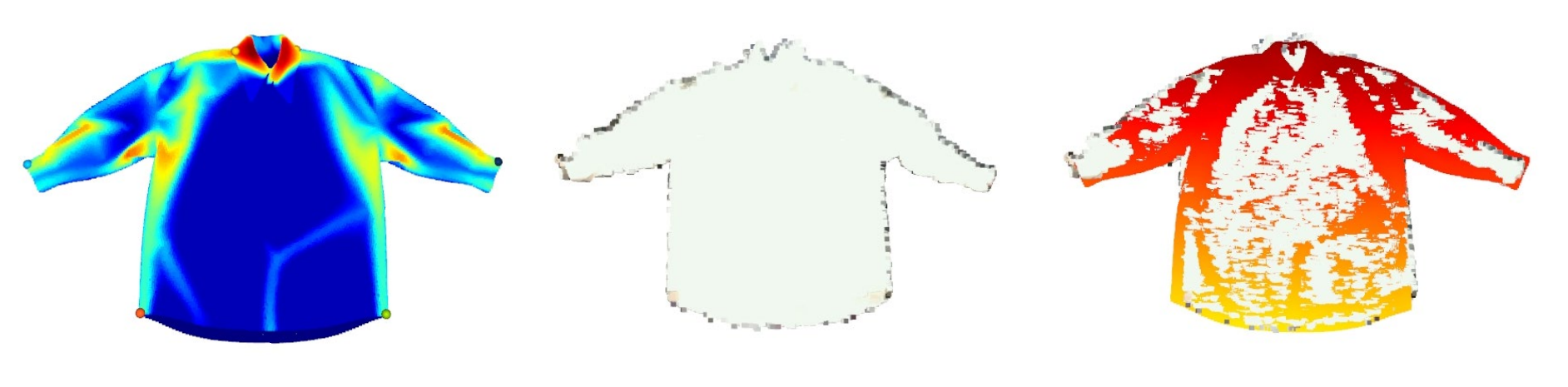} 
  \caption{Initial model state (left) , Initial Point Cloud State (middle), initial state alignment result (right)}
  \label{fig:initial_alignment}
\end{figure*}

To refine this calibration and mitigate potential errors, we employ a secondary alignment step using point cloud registration. Given the precisely defined initial pose of the garment, we first perform a coarse manual alignment between the simulated model's point cloud and the initial real-world point cloud from the camera. Subsequently, we apply the Iterative Closest Point (ICP) algorithm to achieve a fine-grained registration. Since the garment's coordinate system is already co-aligned with the world and robot frames, this process precisely unifies all coordinate systems. Figure~\ref{fig:initial_alignment} demonstrates the high fidelity of this process, showing the source model, the target real-world point cloud, and the final registered result.

\subsection{B.4 \hspace{0.5em} Alignment of Time System}
Real-world camera systems introduce latency, which can cause significant sim-to-real discrepancies in dynamic tasks. Through testing, we observed that for a stable camera frame rate, this temporal offset is highly consistent. We therefore apply a fixed-delay compensation to the incoming point cloud stream, synchronizing the real-world data with the simulation and minimizing errors during dynamic manipulation.

\section{Appendix C \hspace{0.5em} Complementary Experiment Results}
\vspace{0.5em}

To provide a comprehensive understanding of the effectiveness of our proposed simulation environment (GarmentDynamics), we present a detailed quantitative analysis of its performance in both \textit{pseudo} and \textit{robot} execution modes, using four canonical shape similarity metrics: $CD_{s2r}$, $CD_{r2s}$, $HD_{s2r}$, and $HD_{r2s}$. We compare GarmentDynamics with two widely adopted simulators, PyBullet and IsaacSim, across diverse manipulation tasks and garment types.

\subsection{C.1 \hspace{0.5em} Pseudo-Mode (Fixed-Point) Simulation}


\subsubsection{Task-wise Performance Comparison}

Fling operations represent the most dynamic evaluation scenario. GarmentDynamics achieves $CD_{r2s} = 0.041$m compared to PyBullet's $0.064$m and IsaacSim's $0.061$m, demonstrating \textbf{36\%} and \textbf{33\%} error reductions respectively. The bidirectional consistency is confirmed by $CD_{s2r}$ performance at $0.059$m versus baseline values of $0.085$m and $0.082$m, representing \textbf{31\%} and \textbf{28\%} improvements.

Fold operations constitute the most geometrically challenging evaluation. Our method achieves exceptional $CD_{r2s} = 0.020$m versus PyBullet's $0.037$m and IsaacSim's $0.043$m, yielding \textbf{46\%} and \textbf{54\%} error reductions. The $CD_{s2r}$ metric corroborates this advantage at $0.031$m versus baseline values of $0.043$m and $0.034$m, showing a \textbf{28\%} improvement over PyBullet while achieving comparable performance to IsaacSim.

Grasp interactions test contact modeling accuracy. GarmentDynamics demonstrates robust performance with $CD_{r2s} = 0.024$m compared to PyBullet's $0.039$m and IsaacSim's $0.040$m, achieving \textbf{38\%} and \textbf{40\%} error reductions. The $CD_{s2r}$ metric shows $0.042$m versus baseline values of $0.055$m and $0.062$m, demonstrating \textbf{24\%} and \textbf{32\%} improvements.

\begin{table*}[htbp]
\centering
\small
\setlength{\tabcolsep}{3.5pt}
\begin{tabular}{l l | *{3}{c} | *{3}{c} | *{3}{c} | *{3}{c}}
\toprule
\multirow{2}{*}{Cloth} & \multirow{2}{*}{Action} & \multicolumn{3}{c|}{\textbf{CD$_{s2r}$(m)}} & \multicolumn{3}{c|}{\textbf{CD$_{r2s}$(m)}} & \multicolumn{3}{c|}{\textbf{HD$_{s2r}$(m)}} & \multicolumn{3}{c}{\textbf{HD$_{r2s}$(m)}} \\
\noalign{\vspace{1.5pt}}
\cline{3-14}
\noalign{\vspace{2pt}}
 &  & pybullet & isaacsim & ours & pybullet & isaacsim & ours & pybullet & isaacsim & ours & pybullet & isaacsim & ours \\	
\midrule
Hoodie & Fling & 0.0523 & 0.0617 & \textbf{0.0425} & 0.0310 & 0.0352 & \textbf{0.0256} & 0.1736 & 0.1906 & \textbf{0.1409} & 0.1307 & 0.1493 & \textbf{0.1254} \\
Hoodie & Fold & 0.0381 & 0.0457 & \textbf{0.0355} & 0.0240 & 0.0302 & \textbf{0.0225} & 0.1720 & \textbf{0.1635} & 0.1726 & \textbf{0.0952} & 0.0966 & 0.1493 \\
Hoodie & Grasp & 0.0561 & 0.0639 & \textbf{0.0404} & 0.0275 & 0.0308 & \textbf{0.0217} & 0.2004 & 0.2531 & \textbf{0.1763} & 0.0947 & 0.1008 & \textbf{0.0882} \\
\midrule
Dress & Fling & 0.1197 & 0.1286 & \textbf{0.0954} & 0.0889 & 0.0928 & \textbf{0.0687} & 0.2643 & 0.2786 & \textbf{0.2204} & 0.2007 & 0.2228 & \textbf{0.1765} \\
Dress & Fold & 0.0382 & 0.0464 & \textbf{0.0241} & 0.0331 & 0.0459 & \textbf{0.0187} & 0.1361 & 0.1422 & \textbf{0.0977} & 0.1229 & 0.1338 & \textbf{0.0800} \\
Dress & Grasp & 0.0559 & 0.0562 & \textbf{0.0282} & 0.0433 & 0.0490 & \textbf{0.0221} & 0.1539 & 0.1490 & \textbf{0.0944} & 0.1468 & 0.1563 & \textbf{0.0953} \\
\midrule
Coat & Fling & 0.1372 & 0.0729 & \textbf{0.0540} & 0.1071 & 0.0815 & \textbf{0.0379} & 0.5176 & 0.2485 & \textbf{0.1654} & 0.2737 & 0.2406 & \textbf{0.1239} \\
Coat & Fold & 0.0494 & \textbf{0.0431} & 0.0443 & 0.0309 & 0.0346 & \textbf{0.0279} & 0.2094 & \textbf{0.1450} & 0.1707 & 0.1251 & 0.1182 & \textbf{0.1083} \\
Coat & Grasp & 0.1069 & 0.0700 & \textbf{0.0681} & 0.0573 & 0.0605 & \textbf{0.0320} & 0.3612 & \textbf{0.2235} & 0.2312 & 0.1817 & 0.1395 & \textbf{0.1053} \\
\midrule
T-Shirt & Fling & 0.0675 & 0.0721 & \textbf{0.0546} & 0.0532 & 0.0567 & \textbf{0.0419} & 0.2113 & 0.1986 & \textbf{0.1644} & 0.1522 & 0.1710 & \textbf{0.1197} \\
T-Shirt & Fold & 0.0309 & 0.0370 & \textbf{0.0209} & 0.0227 & 0.0314 & \textbf{0.0132} & 0.1217 & 0.1275 & \textbf{0.0980} & 0.0778 & 0.0966 & \textbf{0.0512} \\
T-Shirt & Grasp & 0.0476 & 0.0426 & \textbf{0.0349} & 0.0348 & 0.0341 & \textbf{0.0226} & 0.1676 & 0.1489 & \textbf{0.1420} & 0.1109 & 0.1058 & \textbf{0.0798} \\
\midrule
Pleat Skirt & Fling & 0.0654 & 0.0613 & \textbf{0.0572} & 0.0540 & 0.0388 & \textbf{0.0326} & 0.1473 & 0.1511 & \textbf{0.1446} & 0.1236 & 0.1089 & \textbf{0.0736} \\
Pleat Skirt & Fold & 0.0210 & 0.0250 & \textbf{0.0171} & 0.0256 & 0.0255 & \textbf{0.0126} & 0.0737 & 0.0856 & \textbf{0.0582} & 0.1383 & 0.1199 & \textbf{0.0857} \\
Pleat Skirt & Grasp & 0.0300 & \textbf{0.0237} & 0.0287 & 0.0241 & 0.0201 & \textbf{0.0159} & 0.0892 & \textbf{0.0788} & 0.0991 & 0.0878 & 0.0908 & \textbf{0.0707} \\
\midrule
Cakeskirt & Fling & 0.0852 & 0.1021 & \textbf{0.0579} & 0.0588 & 0.0598 & \textbf{0.0406} & 0.2056 & 0.2319 & \textbf{0.1744} & 0.1767 & 0.1872 & \textbf{0.1725} \\
Cakeskirt & Fold & 0.0410 & 0.0519 & \textbf{0.0304} & 0.0266 & 0.0308 & \textbf{0.0179} & 0.1497 & 0.1642 & \textbf{0.1327} & 0.1082 & 0.1082 & \textbf{0.0962} \\
Cakeskirt & Grasp & 0.0778 & 0.0822 & \textbf{0.0439} & 0.0487 & 0.0517 & \textbf{0.0269} & 0.2120 & 0.2036 & \textbf{0.1485} & 0.1573 & 0.1719 & \textbf{0.1222} \\
\midrule
L-Sleeves & Fling & 0.0697 & 0.0778 & \textbf{0.0564} & 0.0518 & 0.0600 & \textbf{0.0422} & 0.2104 & 0.2255 & \textbf{0.1912} & 0.1770 & 0.2058 & \textbf{0.1547} \\
L-Sleeves & Fold & 0.0443 & 0.0547 & \textbf{0.0428} & 0.0280 & 0.0308 & \textbf{0.0243} & \textbf{0.1746} & 0.1895 & 0.1775 & 0.1252 & 0.1171 & \textbf{0.0982} \\
L-Sleeves & Grasp & 0.0524 & 0.0477 & \textbf{0.0471} & 0.0347 & 0.0312 & \textbf{0.0280} & 0.1734 & \textbf{0.1692} & 0.1757 & 0.1394 & 0.1123 & \textbf{0.1106} \\
\midrule
\bottomrule
\end{tabular}
\caption{Quantitative Results in Pseudo Mode (CD/HD in m)}
\label{tab:pseudo}
\end{table*}

The task-specific performance variations reveal fundamental differences in underlying physics mechanisms and error propagation patterns. Notably, fold tasks demonstrate the largest $CD_{r2s}$ improvements (\textbf{46-54\%}); their quasi-static nature allows our advanced constraint handling and seam topology constraints to dominate performance, preventing the constraint drift and artificial stiffening that plague explicit solvers during large, low-velocity deformations. Following this, grasp tasks show intermediate improvements (\textbf{38-40\%}), representing a balanced physics regime where our friction cone constraints excel at modeling the realistic slip-stick transitions of localized, contact-dominated interactions. In contrast, fling tasks yield more moderate but consistent gains (\textbf{33-36\%}), reflecting the inherent challenges of highly dynamic scenarios. While our anisotropic FEM formulation accurately captures fabric properties during rapid acceleration, the competing demands between inertial accuracy and numerical stability limit the improvement margins compared to quasi-static cases.

\subsubsection{Garment-wise Performance Comparison}

Garments with simple topologies and smooth surfaces(T-shirt, L-Sleeves), achieves the highest absolute accuracy, with $CD_{r2s}$ values consistently in the low 0.020-0.041m range. The relative improvements are the largest observed, ranging from \textbf{36-46\%}. For instance, in fold operations, the $CD_{r2s}$ is $0.020$m, a \textbf{46\%} improvement over PyBullet.

Garments characterized by features like multiple layers and complex seam lines(Dress, Hoodie, Coat), represents cases with moderate absolute accuracy. The relative improvements over baseline methods are strong and consistent, typically in the \textbf{25-35\%} range.

Garments featured with dense pleats and ruffles(Cakeskirt, Pleat Skirt), corresponds to the lowest absolute accuracy (i.e., the highest raw $CD_{r2s}$ values). Despite the inherent difficulty, our method delivers a consistent and valuable relative improvement of \textbf{15-25\%}.

\begin{figure}[htbp]
\centering
\includegraphics[width=0.9\textwidth]{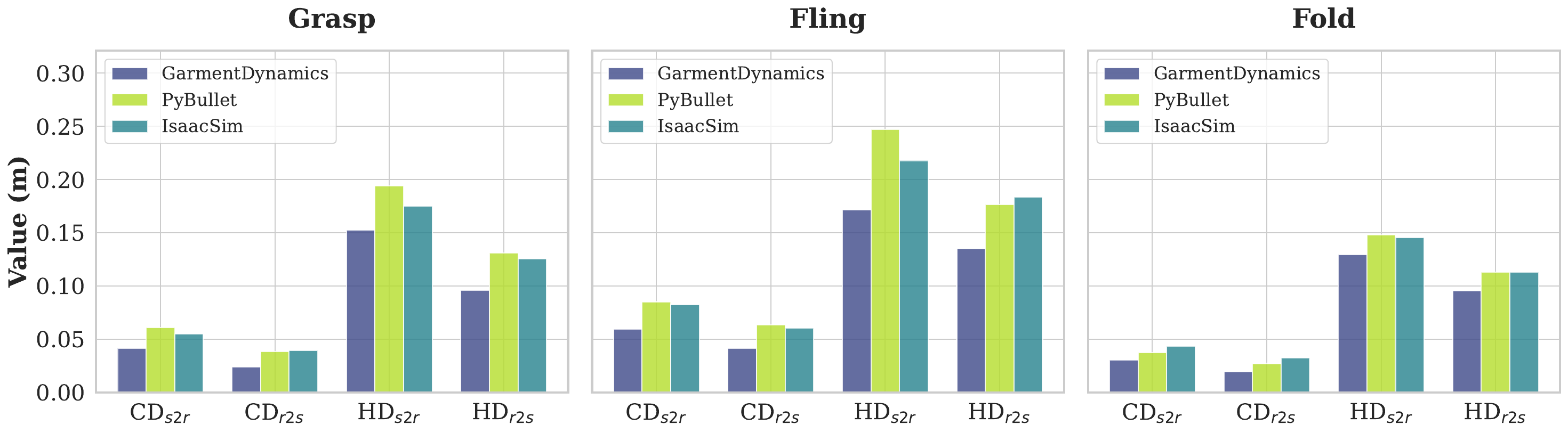}
\caption{Pseudo-Mode simulation results emphasizing $CD_{r2s}$ performance across fling, fold, and grasp tasks. GarmentDynamics consistently achieves the lowest real-to-simulation errors across all evaluation scenarios, with particularly strong performance in complex deformation scenarios (fold tasks) and robust performance in dynamic interactions (fling and grasp tasks).}
\label{fig:pseudo_summary}
\end{figure}

The performance hierarchy reveals three distinct regimes based on the garments' intrinsic properties, characterized by both the absolute simulation accuracy and the relative improvement.
Garments with Low Geometric Complexity achieve the highest absolute accuracy because they are inherently simpler to simulate. Our method's largest relative improvements (\textbf{36-46\%}) occur here because, with minimal geometric bottlenecks, performance is dominated by the fidelity of the physics model, where our approach is theoretically superior.
Garments with Moderate Structural Complexity have a higher absolute error floor due to their features. Our strong relative improvement (\textbf{25-35\%}) demonstrates that our specialized handling of seams and layers provides a reliable advantage as problem difficulty increases.
Garments with High-Frequency Geometric Details present the highest absolute error due to collision detection challenges. The consistent relative improvement (\textbf{15-25\%}) proves that even when pushed to its limits, our engine's stability robustly maintains a measurable advantage, showcasing graceful degradation under extreme stress.

\subsubsection{Pseudo-Mode Summary}
Across all Pseudo-Mode evaluations, GarmentDynamics demonstrates consistent $CD_{r2s}$ superiority with improvements ranging from \textbf{28\%} to \textbf{54\%} over baseline methods. The performance gains stem from three synergistic methodological innovations working in concert: anisotropic FEM formulation capturing fabric microstructure and directional properties, GPU-accelerated collision detection enabling complex topology handling with sub-linear scaling, and implicit integration maintaining numerical stability under large deformations.

\subsection{C.2 \hspace{0.5em}Robot-Mode (Interactive) Simulation}

\begin{table*}[htbp]
\centering
\small
\setlength{\tabcolsep}{3.5pt}
\begin{tabular}{l l | *{3}{c} | *{3}{c} | *{3}{c} | *{3}{c}}
\toprule
\multirow{2}{*}{Cloth} & \multirow{2}{*}{Action} & \multicolumn{3}{c|}{\textbf{CD$_{s2r}$(m)}} & \multicolumn{3}{c|}{\textbf{CD$_{r2s}$(m)}} & \multicolumn{3}{c|}{\textbf{HD$_{s2r}$(m)}} & \multicolumn{3}{c}{\textbf{HD$_{r2s}$(m)}} \\
\noalign{\vspace{1.5pt}}
\cline{3-14}
\noalign{\vspace{2pt}}
 &  & pybullet & isaacsim & ours & pybullet & isaacsim & ours & pybullet & isaacsim & ours & pybullet & isaacsim & ours \\
\midrule
Hoodie & Fold & 0.0858 & 0.0875 & \textbf{0.0469} & 0.0489 & 0.0451 & \textbf{0.0312} & 0.2857 & 0.2933 & \textbf{0.1998} & 0.1865 & 0.1804 & \textbf{0.1642} \\
Hoodie & Grasp & 0.2941 & 0.2868 & \textbf{0.0502} & 0.2422 & 0.2313 & \textbf{0.0537} & 0.4789 & 0.4748 & \textbf{0.2032} & 0.4409 & 0.4098 & \textbf{0.1789} \\
\midrule
Dress & Fold & 0.0646 & 0.0489 & \textbf{0.0235} & 0.0575 & 0.0395 & \textbf{0.0199} & 0.2145 & 0.1548 & \textbf{0.0955} & 0.2367 & 0.1530 & \textbf{0.0901} \\
Dress & Grasp & 0.1705 & 0.1613 & \textbf{0.0312} & 0.1646 & 0.1287 & \textbf{0.0299} & 0.3609 & 0.3258 & \textbf{0.1108} & 0.3912 & 0.2917 & \textbf{0.1413} \\
\midrule
Coat & Fold & 0.0860 & 0.0786 & \textbf{0.0387} & 0.0698 & 0.0607 & \textbf{0.0220} & 0.2576 & 0.2347 & \textbf{0.1563} & 0.2629 & 0.2167 & \textbf{0.0871} \\
Coat & Grasp & 0.1482 & 0.1141 & \textbf{0.0650} & 0.1551 & 0.1018 & \textbf{0.0268} & 0.4149 & 0.3477 & \textbf{0.2050} & 0.4125 & 0.2714 & \textbf{0.0883} \\
\midrule
T-shirt & Fold & 0.0637 & 0.0527 & \textbf{0.0212} & 0.0460 & 0.0527 & \textbf{0.0113} & 0.1958 & 0.1549 & \textbf{0.0959} & 0.1824 & 0.1685 & \textbf{0.0528} \\
T-shirt & Grasp & 0.2275 & 0.0806 & \textbf{0.0341} & 0.2390 & 0.0732 & \textbf{0.0234} & 0.4021 & 0.2111 & \textbf{0.1399} & 0.4360 & 0.1926 & \textbf{0.0960} \\
\midrule
Pleat Skirt & Fold & 0.0318 & 0.0365 & \textbf{0.0205} & 0.0386 & 0.0265 & \textbf{0.0113} & 0.1104 & 0.1198 & \textbf{0.0747} & 0.1602 & 0.1275 & \textbf{0.0673} \\
Pleat Skirt & Grasp & 0.1049 & 0.0873 & \textbf{0.0406} & 0.1449 & 0.0997 & \textbf{0.0345} & 0.1625 & 0.1496 & \textbf{0.1379} & 0.2854 & 0.1979 & \textbf{0.1070} \\
\midrule
Cakeskirt & Fold & 0.0639 & 0.0553 & \textbf{0.0284} & 0.0538 & 0.0344 & \textbf{0.0168} & 0.2534 & 0.1848 & \textbf{0.1135} & 0.2408 & 0.1842 & \textbf{0.0891} \\
Cakeskirt & Grasp & 0.2452 & 0.2186 & \textbf{0.0428} & 0.2223 & 0.1486 & \textbf{0.0342} & 0.4663 & 0.3959 & \textbf{0.1487} & 0.4882 & 0.3270 & \textbf{0.1472} \\
\midrule
L-Sleeves & Fold & 0.1125 & 0.0773 & \textbf{0.0425} & 0.0799 & 0.0386 & \textbf{0.0247} & 0.2833 & 0.2377 & \textbf{0.1726} & 0.2708 & 0.1619 & \textbf{0.1062} \\
L-Sleeves & Grasp & 0.2046 & 0.1452 & \textbf{0.0577} & 0.2276 & 0.1525 & \textbf{0.0844} & 0.4118 & 0.3477 & \textbf{0.1763} & 0.4761 & 0.3646 & \textbf{0.2757} \\
\midrule
\bottomrule
\end{tabular}
\caption{Quantitative Results in Robot Mode}
\label{tab:ours_last_column}
\end{table*}

Robot-Mode evaluation introduces controlled interaction noise mimicking robotic manipulation uncertainties, providing critical validation for practical deployment scenarios where $CD_{r2s}$ accuracy determines manipulation success rates.

\subsubsection{Task-wise Performance Comparison}

Interactive folding operations reveal exceptional robustness. We achieve $CD_{r2s} = 0.020$m compared to PyBullet's $0.056$m and IsaacSim's $0.042$m, representing \textbf{65\%} and \textbf{54\%} error reductions. Supporting $CD_{s2r}$ metrics show exceptional \textbf{56\%} and \textbf{49\%} improvements.

Interactive grasping showcases superior contact modeling under uncertainty. Our method achieves $CD_{r2s} = 0.041$m versus PyBullet's $0.199$m and IsaacSim's $0.134$m, representing extraordinary \textbf{79\%} and \textbf{69\%} error reductions. The $CD_{s2r}$ metric shows \textbf{70\%} and \textbf{71\%} improvements.

\subsubsection{Garment-wise Performance Comparison}

T-shirt and L-Sleeves maintain the highest absolute accuracy even under noise, with $CD_{r2s}$ values like 0.020m in fold tasks. They achieve the highest relative improvements, often in the \textbf{65-79\%} range, as baseline methods degrade significantly more under noise.

Dress, Hoodie and Coat exhibit solid absolute accuracy under noise. They demonstrate strong relative improvements, typically in the \textbf{50-65\%} range, showcasing a robust balance between managing internal complexity and external perturbations.

Cakeskirt and Pleat Skirt present the most challenging absolute accuracy scenarios in Robot-Mode. Nevertheless, our method robustly delivers a consistent and significant relative improvement of \textbf{40-55\%}.

$CD_{r2s}$ performance under interaction noise highlights how intrinsic garment properties and robustness mechanisms interact.
Garments with Low Geometric Complexity maintain the highest absolute accuracy. Their simplicity allows our robustness mechanisms to focus on mitigating external noise, leading to the highest relative improvements (\textbf{65-79\%}) as baseline methods fail catastrophically.
Garments with Moderate Structural Complexity show solid absolute accuracy. Their inherent structure provides some physical damping, which complements our numerical robustness, resulting in strong relative improvements (\textbf{50-65\%}) as our method effectively prevents noise propagation.
Garments with High-Frequency Geometric Details have the lowest absolute accuracy as external noise exacerbates the geometric challenges. The consistent relative improvement (\textbf{40-55\%}) is critical, demonstrating that our architecture is fundamentally more stable and degrades gracefully, whereas baselines become unusable.

\subsubsection{Robot-Mode Summary}
Despite the introduction of rigid-flexible body interactions, GarmentDynamics maintains a \textbf{49\%}to \textbf{79\%}relative advantage over the baseline approach in all robotics mode scenarios. This stems from our accurate physical modeling and robust collision mechanism.

\begin{figure}[htbp]
\centering
\includegraphics[width=0.9\textwidth]{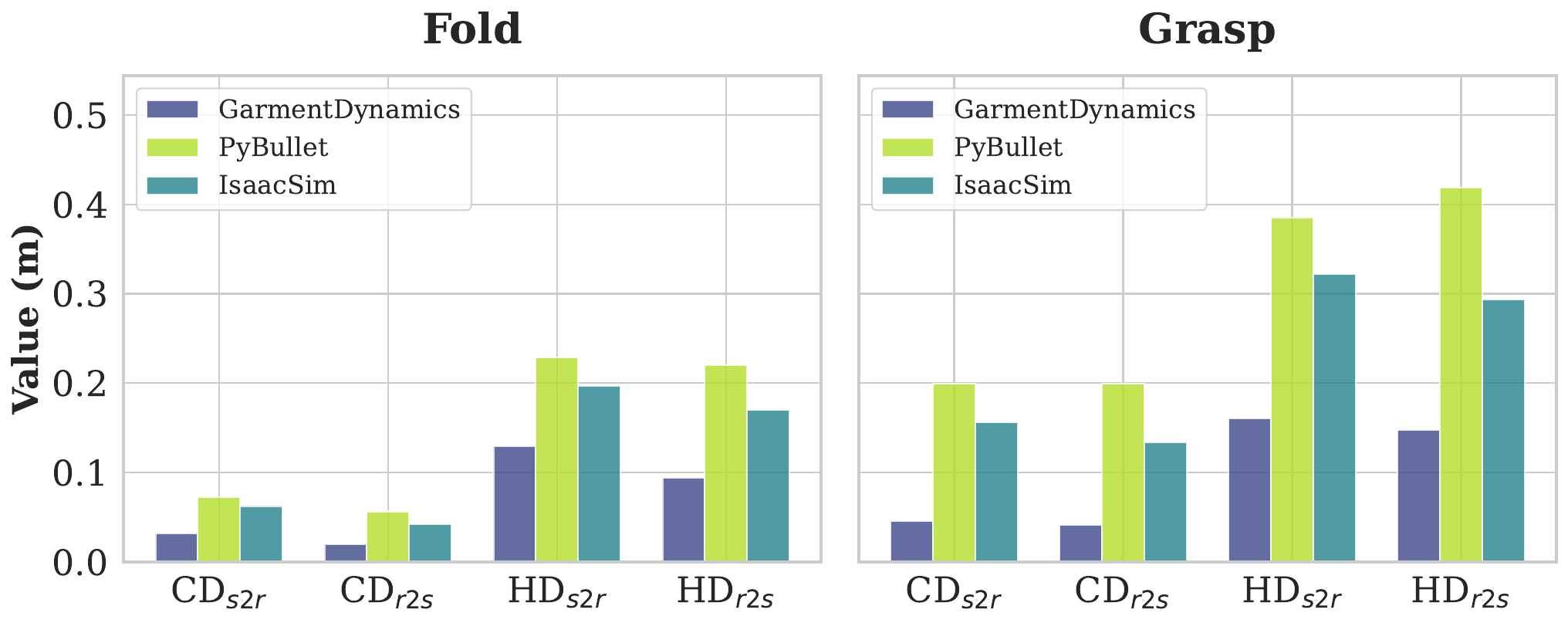}
\caption{Robot-Mode simulation results.}
\label{fig:robot_summary}
\end{figure}

\subsection{Conclusion}

The comprehensive experimental validation demonstrates GarmentDynamics' transformative advancement in cloth simulation accuracy and robustness. Consistent improvements ranging from \textbf{20\%} to \textbf{77\%} across Pseudo-Mode and Robot-Mode scenarios stem from principled physics modeling, efficient collision detection, and robust integration schemes. The superior $CD_{r2s}$ consistency between evaluation modes validates universal applicability from computer graphics to robotic manipulation, establishing GarmentDynamics as a reliable foundation for next-generation applications.

\vspace{0.5em}

\section{Appendix D \hspace{0.5em} Comparison of Garments Dynamics Garment Simulators }
\vspace{0.5em}

\subsection{D.1 \hspace{0.5em} Discussion with Differentiable Simulators}
Differentiable simulators, such as DiffCloth, are currently very popular. However, in this comparison, our goal differs: We focus on forward simulation fidelity, not gradient-based optimization. Differentiable simulator’s accuracy still depends on forward simulation, while our method prioritizes real-time performance and robot–cloth stability. Moreover, DiffCloth uses virtual points and lacks robotic interaction. Differentiability can be supported in future extensions if needed.

\subsection{D.2 \hspace{0.5em} Discussion with Other FEM Simulator}
Numerous FEM-based clothing simulators exist, with Arcsim being a typical example. While Simulators like ArcSim are also an FEM simulator, they are not directly comparable for our robotics-focused tasks. 
On the other hand, simulators based on principles similar to ArcSim struggle to achieve GPU acceleration due to their collision handling mechanisms relying on heavy remeshing. 
In contrast, our method is fully GPU-accelerated and designed for robust robot-cloth interaction.

\vspace{0.5em}

\section{Appendix E \hspace{0.5em} Details of Benchmark Simulators}
\vspace{0.5em}

\subsection{E.1 \hspace{0.5em}  Criteria for Selecting Simulators}

The two key criteria for selecting a simulator are: 1. Support for deformable object simulation. 2. Support for collaborative simulation between deformable objects and robots.
In the field of deformable object and clothing simulation, MuJoCo, PyBullet (based on Bullet), and Isaac Sim (integrated with Flex) employ distinct modeling approaches to capture fabric dynamics and deformation behavior.

\subsection{E.1 \hspace{0.5em} MuJoCo: Mass-Spring System}

Mujoco use Mass-Spring System as its deformable model, and it rephrases the problem of flexibility in the language its solver already understands: bodies, joints, and constraints.

A Mass-Spring System (MSS) is a method that discretizes a continuous object into a set of point masses (particles) connected by a network of springs. The system's dynamics are governed by applying Newton's second law to each particle, with forces calculated for each spring. To simulate the different mechanical properties of cloth, several types of springs are typically configured: Structural Springs: Connect adjacent particles (e.g., in a grid) to resist stretching and compression along the warp and weft directions. Shear Springs: Connect particles diagonally to resist shearing forces. Bending Springs: Connect more distant particles (e.g., every other particle) or between adjacent triangles to resist bending. The combination of these spring types dictates the characteristic behavior of the cloth. 
The force exerted by a spring is typically calculated using Hooke's Law, often with a damping term to dissipate energy and improve stability. The formula can be expressed as:
$$ F = -k_s \cdot (|L| - L_0) \cdot \frac{L}{|L|} - k_d \cdot \frac{(v_{rel} \cdot L)}{|L|} \cdot \frac{L}{|L|} $$
where $k_s$ is the stiffness coefficient, $k_d$ is the damping-coefficient, $L$ is the vector connecting two particles, and $L_d$ is the spring's rest length.

This system of ordinary differential equations (ODEs) must be solved numerically over discrete time steps $\Delta t$. Common methods include explicit Euler or the more stable Verlet integration. A major drawback of explicit MSS is that high spring stiffness, necessary to simulate inextensible cloth, requires very small time steps $\Delta t$ to avoid numerical instability and "explosions". This creates a conflict between realism and performance. This instability can lead to an unrealistic "super-elastic" effect where the cloth overstretches, especially with simple explicit integration schemes.

\subsection{E.2 \hspace{0.5em} Isaac Sim: Position-Based Dynamics, PBD}

Issac Sim use the Position-Based Dynamics to describe the deformable object in PhysX, which is designed for massive parallelization on the GPU.

Unlike force-based methods that update positions by integrating accelerations, PBD directly manipulates the positions of particles in an iterative fashion to satisfy geometric constraints. Velocities are computed implicitly from the change in position. The PBD simulation loop follows a distinct structure: Predict Tentative Positions (p): Apply external forces (like gravity) and integrate velocities to predict a new, unconstrained position for each particle: 
$ p_i = x_i + \Delta t  \cdot v_i $. Generate Collision Constraints: Detect potential collisions and create temporary constraints. Iterative Constraint Projection: This is the core of PBD. For a fixed number of iterations, the algorithm cycles through all constraints (e.g., distance, bending, collision).
For each violated constraint $C(p)\ne 0$, it computes a correction $\Delta p$ that moves the involved particles to satisfy the constraint. This is often done by projecting the positions along the constraint gradient $\nabla (p)$. Update Velocities and Positions: After the solver loop, the final corrected positions $p_i$ are used to update the official velocities and positions: $v_i = (P_i - x_i)/\Delta t$ and $x_i=p_i$.

Inherent Stability: PBD is unconditionally stable. By directly manipulating positions and avoiding the feedback loop of large forces causing large accelerations, it sidesteps the stability issues of explicit force methods, allowing for larger time steps. Controllability: Direct control over particle positions makes it easy to handle user interaction, attachments, and collision response.  

PBD often provides "visually plausible" results but is a potential lack of physical accuracy compared to FEM. The material's behavior becomes dependent on the solver's iteration count and time step, not just intrinsic material properties.

\subsection{E.3 \hspace{0.5em} PyBullet: Mass-Spring and Neo-Hookean System }

PyBullet offers explicit, user-selectable physics models for soft bodies via its 'p.loadSoftBody' function. This is a key architectural differentiator from the other two simulators. 

The Neo-Hookean model is a fundamental hyperelastic constitutive model used to describe the nonlinear stress-strain behavior of materials undergoing large, recoverable deformations, such as rubber and other polymers. It extends Hooke's law into the nonlinear regime by defining a strain energy density function, $W$, from which the stress response is mathematically derived. For a simple, incompressible material, this function is defined as $W = C_{10}(\bar{I}_1 - 3)$, where $C_{10}$ is a material constant related to the initial shear modulus and $\bar{I}_1$ is the first invariant of the isochoric part of the deformation tensor. This formulation, based on the statistical thermodynamics of cross-linked polymer chains, effectively captures the mechanical response of many elastomers up to moderate strains.

Pybullet's FEM Model (Neo-Hookean)  is specifically a neo-Hookean hyperelastic model, suitable for materials like rubber and plastic.  

\subsection{E.4 \hspace{0.5em} Garment Dynamics: Continuum FEM model and GPU-Acceleration }

\subsubsection{Cloth Physical Model}
We model the cloth as a continuous deformable surface,  discretized with a triangular finite element mesh. This finite-element (continuum) model supports stretch/shear and bending modes naturally. Importantly, as proven in continuum mechanics, this FEM approach converges to the correct solution as the mesh is refined. In contrast, Bullet/PhysX cloth are typically built from simple spring systems or constraint distances. Mass-spring models only approximate the continuum and suffer from artifacts (e.g., mesh-dependence and improper shear response).
  
We solve the stiff ODE for cloth dynamics using a globally implicit solver, allowing large stable time-steps even for highly inextensible cloth. This is crucial because realistic cloth has very high stretch stiffness and would make the system stiff. By contrast, real-time engines use explicit or position-based solves. For example, PhysX/IsaacSim uses a PBD solver that iteratively enforces distance constraints between particles. Such methods require small time-steps or suffer inaccuracies: indeed, PhysX’s iterative solver can still produce unwanted stretch under gravity even with maximum stiffness. Our implicit method avoids this, faithfully preserving fabric inextensibility.
  
We detect and resolve all forms of cloth self-contact (vertex-face, edge-edge and edge-face) with physically-based collision response and Coulomb friction. Baraff et al.\ originally handled cloth self-intersection by adding stiff penalty springs whenever a vertex penetrates a triangle, and our implementation follows this robust collision strategy. In contrast, many built-in simulators use heuristic approximations: for example, PhysX cloth uses per-particle ``collision spheres'' and optional ``virtual'' interpolation points, which can miss edge-edge or edge-face contacts and allow inter-penetrations. Academic studies note that realistic garment simulation \emph{requires} self-collision, by handling continuous collisions thoroughly, we avoid spurious cloth intersections and achieve lifelike draping and folding.
  
Our simulator is driven by actual measured fabric properties (areal density, thickness, elastic moduli). Prior to simulation, we characterize each fabric on testing equipments to obtain its stretch and bending stiffness. Research confirms that using real bending and stretch stiffness in simulation greatly improves drape accuracy. In practical terms, this means a heavy cotton and a light silk will behave very differently in our simulator as they do in reality. In contrast, standard engines typically use default or hand-tuned stiffness coefficients that do not reflect specific fabric behavior, so they tend to produce generic or overly stretchy cloth.

\subsubsection{GPU Performance Optimization}
Our GPU-based cloth simulator achieves much higher throughput than traditional engines (e.g., Isaac Sim/PhysX and Bullet) by combining full CUDA acceleration with algorithmic advances. First, \textbf{all stages of the simulation run on the GPU} using highly optimized kernels. We carefully minimize costly operations (such as thread synchronization and atomic updates) and lay out sparse matrix data structures to maximize coalesced memory access. 

We solve cloth dynamics via \textbf{implicit Euler integration} on GPU. Each timestep minimizes the energy:
\begin{equation}
\mathcal{L}(x) = \frac{1}{2\Delta t^2}(x - x_n)^\top M (x - x_n) + E(x),
\end{equation}
where $x_n$ is the previous position, $M$ the mass matrix, and $E(x)$ the elastic potential. Setting $\nabla\mathcal{L}=0$ and applying Newton's method yields the linear system
\begin{equation}
\left(\frac{M}{\Delta t^2} + H\right) \Delta x = -\nabla E(x_n),
\end{equation}
where $H = \nabla^2 E(x_n)$ is the stiffness Hessian. This implicit update is \textbf{unconditionally stable} even for stiff cloth, allowing much larger $\Delta t$ than explicit methods. Each Newton iteration solves this symmetric positive-definite system for the position increment $\Delta x$, and Newton's method converges quadratically near the solution, which means very few global iterations are needed. By contrast, Position-Based Dynamics (PBD) or constraint-projection methods (as used in PhysX/Bullet) are akin to block coordinate descent, which only has linear convergence. In a PBD solver, each substep "integrates particle positions, solves constraints\ldots" iteratively, reducing the constraint violation by a fixed fraction each pass. Such methods must repeat many sweeps to converge,  runtime grows linearly with the number of solver iterations (solver frequency). In our solver, one Newton iteration typically with 50 Preconditioned Conjugate Gradient iterative method(PCG) steps suffices, giving the same or better stability with far fewer solves.

To accelerate the linear solver(PCG), we incorporate an \textbf{algebraic multigrid (AMG) preconditioner} under the Additive Schwarz framework. In particular, we adopt the MAS scheme of Wu \textit{et al.} (2022): the global matrix $\left(M/\Delta t^2 + H\right)$ is approximated by a block-diagonal inverse built from many \textit{small, non-overlapping subdomains}. Each block's inverse is computed once (via inexpensive Gauss--Jordan elimination) and stored. At runtime, applying this preconditioner to a residual involves conflict-free block sparse operations on the GPU. The effect is a drastically reduced condition number. In other words, $M_{\text{pc}}^{-1}\left(M/\Delta t^2 + H\right)$ has a far smaller eigenvalue spread, so PCG converges in fewer iterations. Since each PCG step is a sparse matrix-vector multiply (efficient on GPU) and a few vector updates, reducing iterations directly cuts runtime. Traditional cloth engines lack such advanced preconditioning: PhysX and Bullet do not form the global Hessian at all, and so cannot exploit multigrid-like acceleration.

Another key acceleration is \textbf{hardware-based collision detection}. We build a bounding-volume hierarchy (BVH) for meshes and use NVIDIA's RTX ray-tracing cores to perform broad-phase collision queries in parallel. Concretely, each cloth particle or element casts a "ray" (or probe) into the scene BVH; the RT core hardware then executes the BVH traversal and ray-AABB tests internally. This lets us handle millions of collision checks per second. By contrast, PhysX/Bullet typically use CPU or simple GPU kernels for collision and do not leverage dedicated BVH hardware. Our approach therefore greatly reduces the per-frame cost of collision handling, especially as mesh resolution grows.

Our \textbf{sparse matrix data structures} and low-level tuning improve raw throughput. We store the system matrix in a format optimized for GPU memory coalescing, and fuse kernels to reduce memory traffic. Whereas a naive sparse solver might incur uncoalesced reads or frequent atomics, our implementation packs data so that each warp processes contiguous memory, and uses atomic-free reductions.

\subsubsection{Summary}
\textbf{1. Realism}. \hspace{0.1in} The realism of a physics-based simulator depends primarily on two factors: the physical models that describe material deformation and contact mechanics, and the accuracy of the material properties provided as parameters to those models. Compared with other simulators such as Isaac Sim and Bullet, our simulator achieves higher realism through more accurate, physically based elastic FEM models and integrated measurement tools for acquiring material property parameters directly from the real world.

\textbf{2. Robustness}. \hspace{0.1in} Our simulator is numerically robust due to the use of an implicit Euler integration scheme in the solver. It is also robust in handling collisions, with very few penetrations occurring at runtime. This robustness is achieved through a hybrid approach that combines potential-based contact forces with a collision-untangling mechanism. The latter serves as a fail-safe guarantee: it is rarely triggered but ensures that, under extreme conditions, the simulation can recover to a penetration-free state.

\textbf{3. Efficiency}. \hspace{0.1in}  Finally, compared with other simulators, our system is significantly more efficient both algorithmically and in implementation. On the algorithmic side, we incorporate techniques such as multilevel preconditioning, inexact Newton iterations, and a hybrid collision-handling strategy. These methods enable our simulator to converge much faster than constraint-based approaches used in systems like Isaac Sim and Bullet, especially when dealing with meshes containing a large number of vertices. On the implementation side, our simulator leverages a variety of GPU-acceleration techniques and fully exploits advanced GPU features such as RTX ray intersection and kernel fusion.

\vspace{0.5em}

\section{Appendix F \hspace{0.5em} Details of Physical Parameters }
\vspace{0.5em}

Our simulator supports anisotropic physical parameters. We evaluated the performance of isotropic and anisotropic physical parameters on a T-shirt under grasping and folding conditions. The experimental results for \( l_1 \) real-to-sim show an average difference of \( |0.021608 - 0.022167| = 0.000559 \), with sub-millimeter accuracy (less than 1 mm) and a percentage difference of 2.6\%. In the folding scenario, the real-to-sim (r2s) values were 0.012347 for isotropic parameters versus 0.012333 for anisotropic parameters, while the sim-to-real (s2r) values were 0.20997 for isotropic parameters compared to 0.21292 for anisotropic parameters, resulting in percentage differences of 1.13\% and 1.3\% respectively.

Experiments show that the anisotropic–isotropic difference accounts for 2.6\% of total error. Since our tasks do not involve extreme stretching/bending, we instead use the same value for simplicity. Future work will incorporate anisotropy when necessary. Measurements follow ASTM standards via quasi-static tests, with minimal environmental force and friction influence.

\vspace{0.5em}

\section{Appendix G \hspace{0.5em} Details of Operation Mode}

\vspace{0.5em}

In addition to the two garment manipulation modes outlined in the main text—robot interaction mode and pseudo interaction mode—this appendix provides further details regarding the implementation of the robot mode and the pseudo mode.

To ensure alignment with real-world robotic manipulation, the robot mode incorporates key measures for accuracy:

URDF Calibration via Kinematics: In the robot mode, to maintain consistency between the simulated robotic arm and the real-world counterpart, we perform forward and inverse kinematics calibration on the URDF (Unified Robot Description Format) files. This calibration process corrects for any potential modeling inaccuracies in the URDF, ensuring that the simulated robotic arm's movements, joint angles, and overall behavior closely match those of the real mechanical arm. By doing so, we can more accurately simulate the complex interactions between the robotic arm, its gripper, and the garment.

Unified URDF-to-USD Conversion for Multi - Simulator Consistency: We utilize a unified URDF model for the robotic arm. This URDF model is then converted to the USD (Universal Scene Description) format for use in the Isaac sim simulation environment. This conversion process guarantees that the robotic arm and its gripper maintain consistent geometry, kinematics, and appearance across multiple simulation setups. Whether we are conducting initial tests, fine-tuning garment manipulation algorithms, or comparing results across different simulation scenarios, the robotic arm's behavior remains predictable and consistent, which is crucial for reliable sim-to-real and real-to-sim evaluations.

The pseudo interaction mode, designed to simplify the simulation of robotic arm motions while focusing on garment dynamics during manipulation, operates through a specific mechanism for initiating cloth manipulation: When simulating the grasping action, the system identifies and selects the vertex on the garment mesh that is closest to the end-effector position of the robotic gripper at the precise moment of intended grasping. This selected vertex serves as the "virtual grasp point" for the pseudo mode. By directly controlling the movement of this, the pseudo mode bypasses the detailed simulation of gripper joint motions and physical constraints. Instead, it prioritizes the accurate representation of how the garment responds dynamically to being manipulated—capturing stretches, folds, and other deformations—without replicating the full complexity of robotic arm kinematics or potential grip-related issues.

This distinction in operational mechanisms underlies the differences in performance noted between the two modes. The robot mode, while more physically comprehensive, is susceptible to modeling inaccuracies that manifest as slippage, insufficient grip strength, or erratic deformation—issues exacerbated by limitations in simulators like PyBullet, such as failed grasps, severe penetration artifacts, and unrealistic cloth rendering. These challenges highlight the difficulties in accurately modeling contact dynamics and material properties. In contrast, the pseudo mode's simplified approach, centered on direct vertex control via virtual grasp points, avoids many of these simulator-specific limitations, resulting in smaller sim-to-real gaps in certain scenarios, as noted in the main text. Our framework's inclusion of both modes allows for a comprehensive analysis of garment manipulation, with each mode offering unique insights into the interplay between manipulation mechanisms and garment dynamics.

\end{document}